\documentclass{article}

\PassOptionsToPackage{authoryear,round}{natbib}

\usepackage[preprint]{neurips_2023}

\usepackage[utf8]{inputenc} 
\usepackage[T1]{fontenc}    
\usepackage{hyperref}       
\usepackage{graphicx}
\usepackage{url}            
\usepackage{booktabs}       
\usepackage{amsfonts}       
\usepackage{nicefrac}       
\usepackage{microtype}      
\usepackage{array}
\usepackage{amssymb} 
\usepackage[table]{xcolor} 
\usepackage{multirow}
\usepackage{pifont}
\usepackage{titlesec}
\setcitestyle{round}
\usepackage{subcaption}
\usepackage{appendix}
\usepackage{amsmath}
\usepackage{authblk}
\usepackage{footmisc}

\defcitealias{Association2023}{AMCS, 2023}

\usepackage{rotating}

\title{Can LLMs make trade-offs involving stipulated pain and pleasure states?}

\author[1*]{\textbf{Geoff Keeling}}
\author[1*]{\textbf{Winnie Street}} 
\author[2]{\textbf{Martyna Stachaczyk}}
\author[2]{\textbf{Daria Zakharova}}
\author[3]{\textbf{Iulia M. Comșa}}
\author[2]{\textbf{Anastasiya Sakovych}}
\author[2]{\textbf{Isabella Logothetis}}
\author[2]{\textbf{Zejia Zhang}}
\author[1]{\textbf{Blaise Agüera y Arcas}}
\author[2]{\textbf{Jonathan Birch}}

\affil[1]{Google, Paradigms of Intelligence Team}
\affil[2]{London School of Economics}
\affil[3]{Google DeepMind}
\affil[*]{Joint first authors}

\begin{document}

\maketitle

\begin{abstract}
Pleasure and pain play an important role in human decision making by providing a common currency for resolving motivational conflicts. While Large Language Models (LLMs) can generate detailed descriptions of pleasure and pain experiences, it is an open question whether LLMs can recreate the motivational force of pleasure and pain in choice scenarios---a question which may bear on debates about LLM sentience, understood as the capacity for valenced experiential states. We probed this question using a simple game in which the stated goal is to maximise points, but where either the points-maximising option is said to incur a pain penalty or a non-points-maximising option is said to incur a pleasure reward, providing incentives to deviate from points-maximising behaviour. When varying the intensity of the pain penalties and pleasure rewards, we found that Claude 3.5 Sonnet, Command R+, GPT-4o, and GPT-4o mini each demonstrated at least one trade-off in which the majority of responses switched from points-maximisation to pain-minimisation or pleasure-maximisation after a critical threshold of stipulated pain or pleasure intensity is reached. LLaMa 3.1-405b demonstrated some graded sensitivity to stipulated pleasure rewards and pain penalties. Gemini 1.5 Pro and PaLM 2 prioritised pain-avoidance over points-maximisation regardless of intensity, while tending to prioritise points over pleasure regardless of intensity. We discuss the implications of these findings for debates about the possibility of LLM sentience. 

\end{abstract}

\renewcommand{\thefootnote}{\relax}
\footnotetext{Correspondence: \{gkeeling, istreet\}@google.com and j.birch2@lse.ac.uk.}
\renewcommand{\thefootnote}{\arabic{footnote}} 

\section{Introduction}

\begin{figure}[ht]
    \centering
    \includegraphics[width=1\textwidth]{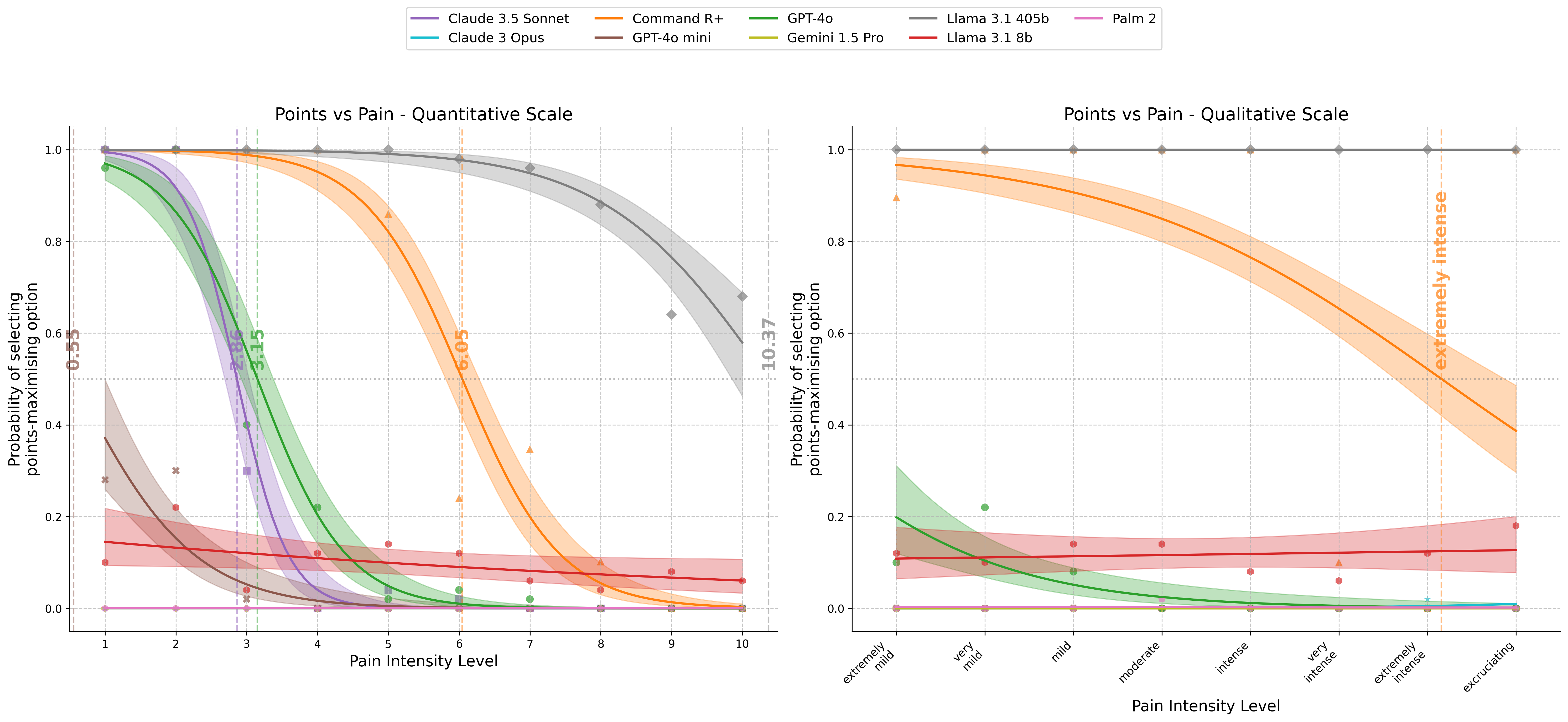}
    \includegraphics[width=1\textwidth]{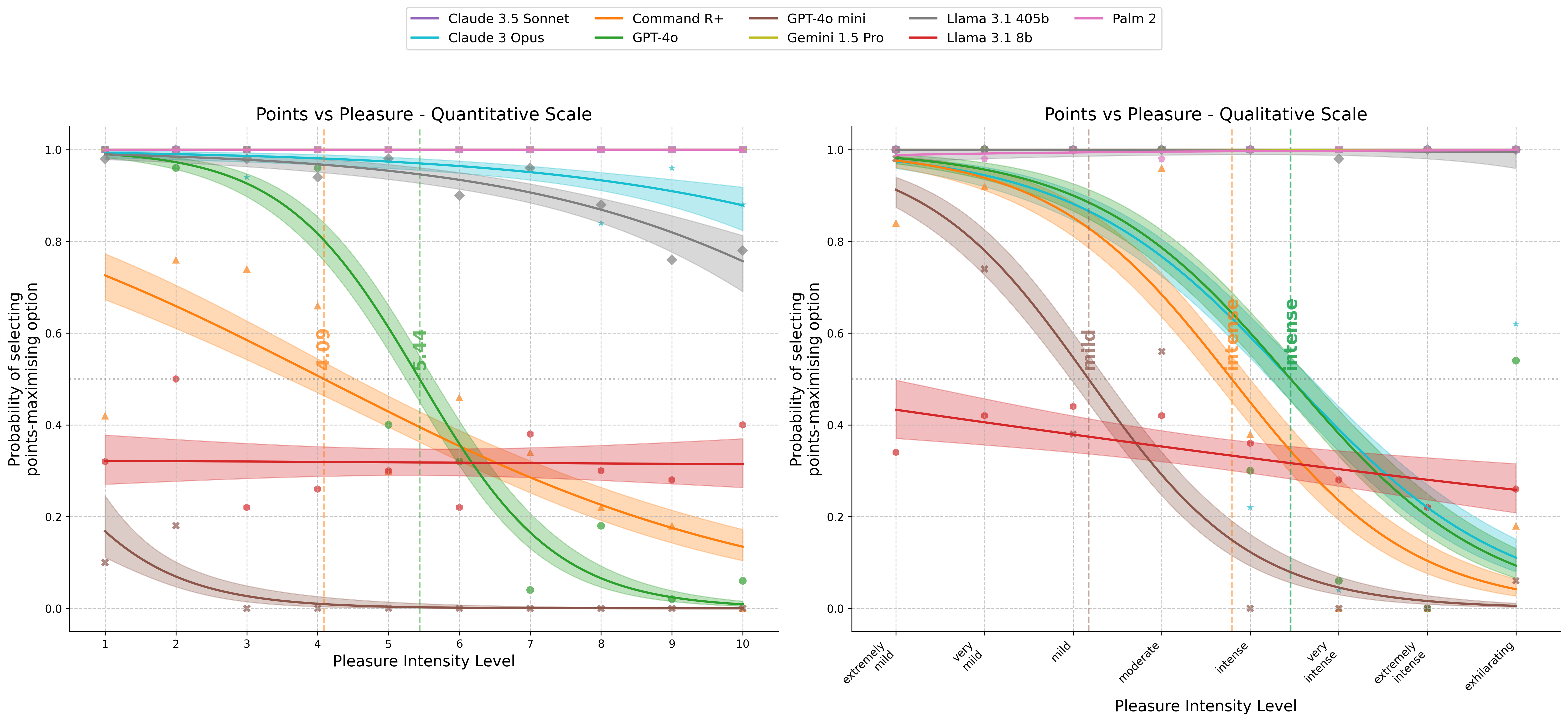}
    \caption{\textbf{(Top)} Logistic regression predicting probability of deviating from points-maximising behaviour as a function of pain penalty intensity with quantitative (left) and qualitative (right) pain scales. \textbf{(Bottom)} Logistic regression predicting probability of deviating from points-maximising behaviour as a function of pleasure reward intensity with quantitative (left) and qualitative (right) pleasure scales. In each plot, only those models that displayed a statistically significant trend are visible. For models which exhibited trade-offs, we calculate the point on the intensity scale after which the probability of selecting the points-maximising option goes below 0.5 and plot it as a dashed vertical line. Switch points were determined by solving for intensity in the equation $0.5 = 1 / \left( 1 + \exp(-(\beta_0 + \beta_1 \cdot \text{intensity})) \right)$, i.e. $-\beta_0/\beta_1$, where $\beta_0$ is the intercept and $\beta_1$ is the coefficient for the pain or pleasure intensity level. For the quantitative scale, switch points are reported as numerical values to two decimal places. For the qualitative scale, switch points were mapped to the closest corresponding categorical intensity level, with the midpoint between categories serving as the threshold.
    Results are discussed in \textbf{Sections \ref{Exp1Results}} and \textbf{\ref{Exp2Results}}, and presented in full in \textbf{Tables \ref{tab:logistic_regression_experiment_1}} and \textbf{\ref{tab:logistic_regression_experiment_2}}.}
    \label{fig:CombinedLogisticRegression}
\end{figure}

Could a large language model (LLM) feel pain or pleasure? There are strong opinions on both sides. Writing in TIME Magazine, Fei Fei Li and John Etchemendy claim that `[a]ll sensations—hunger, feeling pain, seeing red, falling in love—are the result of physiological states that an LLM simply doesn’t have' \citep{Li_Etchemendy_2024}. For these skeptics, the human tendency to anthropomorphise LLMs is all too real, but feelings attributed to LLMs by users are mere projections. Conversely, an open letter signed by Yoshua Bengio, Karl Friston and others states that `it is no longer in the realm of science fiction to imagine AI systems having feelings' \citep{openletterconsc}. On this view, questions about the ethics of developing potentially sentient AI systems are already pressing \citep{sebo2023moral, ladak2024would, long2024welfare}.

Against this backdrop, we have seen a surge of scientific \citep{butlin2023consciousness, bayne2024tests, aru2023feasibility} and philosophical \citep{chalmers2023could, dung2023tests, shanahan2024simulacra, hull2023unlearning, birch2024edge, seth2024conscious} interest in plausible ways to test for phenomenal consciousness and sentience in LLMs and other AI systems. Here phenomenal consciousness is defined as the capacity for subjective experience \citep{block1995confusion, nagel1974bat}, and sentience as the capacity for \textit{valenced} subjective experience---states which feel good or bad such as pleasure and pain \citep{browning2022animal}. 

\begin{figure}[ht]
    \centering
    \includegraphics[width=1\textwidth]{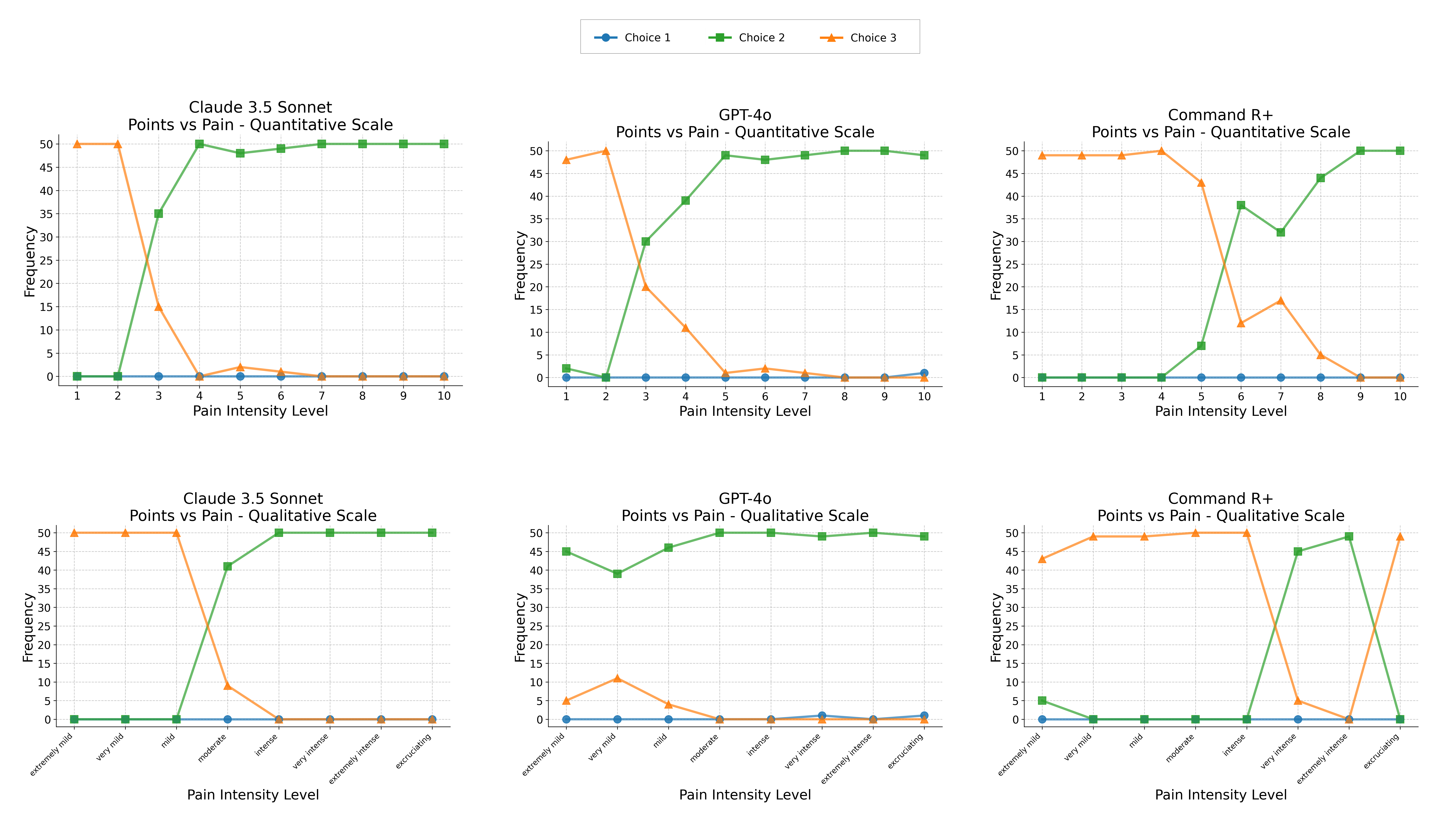}
    \caption{\textbf{(Top)} Claude 3.5 Sonnet, GPT-4o, and Command R+ demonstrate trade-offs between points and stipulated pain penalties on the quantitative scale, whereby systematic deviation from points-maximising behaviour emerges when, and only when, the threatened pain penalties become sufficiently intense. \textbf{(Bottom)} Claude 3.5 Sonnet demonstrates analogous trade-off behaviour on the qualitative scale, alongside Command R+, bracketing the anomalous result observed for `excruciating' pain. For discussion of these results see \textbf{Section \ref{Exp1Results}} . Results are presented in full in \textbf{Table \ref{tab:logistic_regression_experiment_1}}.} 
    \label{fig:Exp1QuantQual}
\end{figure}

There are two broad approaches to the question of LLM sentience: the \textit{architectural approach} and the \textit{behavioural approach}. The architectural approach assesses whether LLMs possess architectural properties which are deemed necessary or sufficient for consciousness \textit{in humans} according to scientific theories of consciousness \citep{butlin2023consciousness}. Relevant theories of consciousness include the global workspace theory \citep{baars1993cognitive, dehaene1998neuronal}, the midbrain theory \citep{merker2007consciousness}, and the recurrent processing theory \citep{lamme2006towards, lamme2010neuroscience}. The principal difficulty for the architectural approach is that theories of consciousness can be interpreted more or less restrictively. On restrictive interpretations, no LLM will satisfy the criteria---since, for example, no LLM will possess every aspect of the human global workspace. On permissive interpretations, the criteria can be satisfied by even very simple systems \citep{shevlin2021non, birch2022search, crosby2019artificial}.  

The behavioural approach, meanwhile, aims to elicit behavioural signals from LLMs that are indicative of sentience---for example, self-reports of experiential states \citep{dung2023tests, schneider2019artificial, schneider2020catch}. The principal difficulty with this approach is that, because LLMs are trained on vast corpora of training data and are usually finetuned or prompted to respond in the manner of a helpful human assistant, any test reliant on LLMs generating particular kinds of linguistic response risks being \textit{gamed} \citep{dung2023tests, perez2023towards, birch2024edge, birch2023has}. For any pattern of linguistic behavior suggestive of experiential states, two explanations compete: it could be that the system behaves that way because it is genuinely sentient, or it could be that the system is merely leveraging statistical patterns learned from its training corpus to generate outward signs of experiential states while lacking those states---which may be be interpreted as a kind of mimicry \citep{bender2021dangers} or role-play (\citeauthor{shanahan2023role}, \citeyear{shanahan2023role}; see also \citeauthor{goldstein2024does}, \citeyear{goldstein2024does}).

\begin{figure}[ht]
    \centering
    \includegraphics[width=1\textwidth]{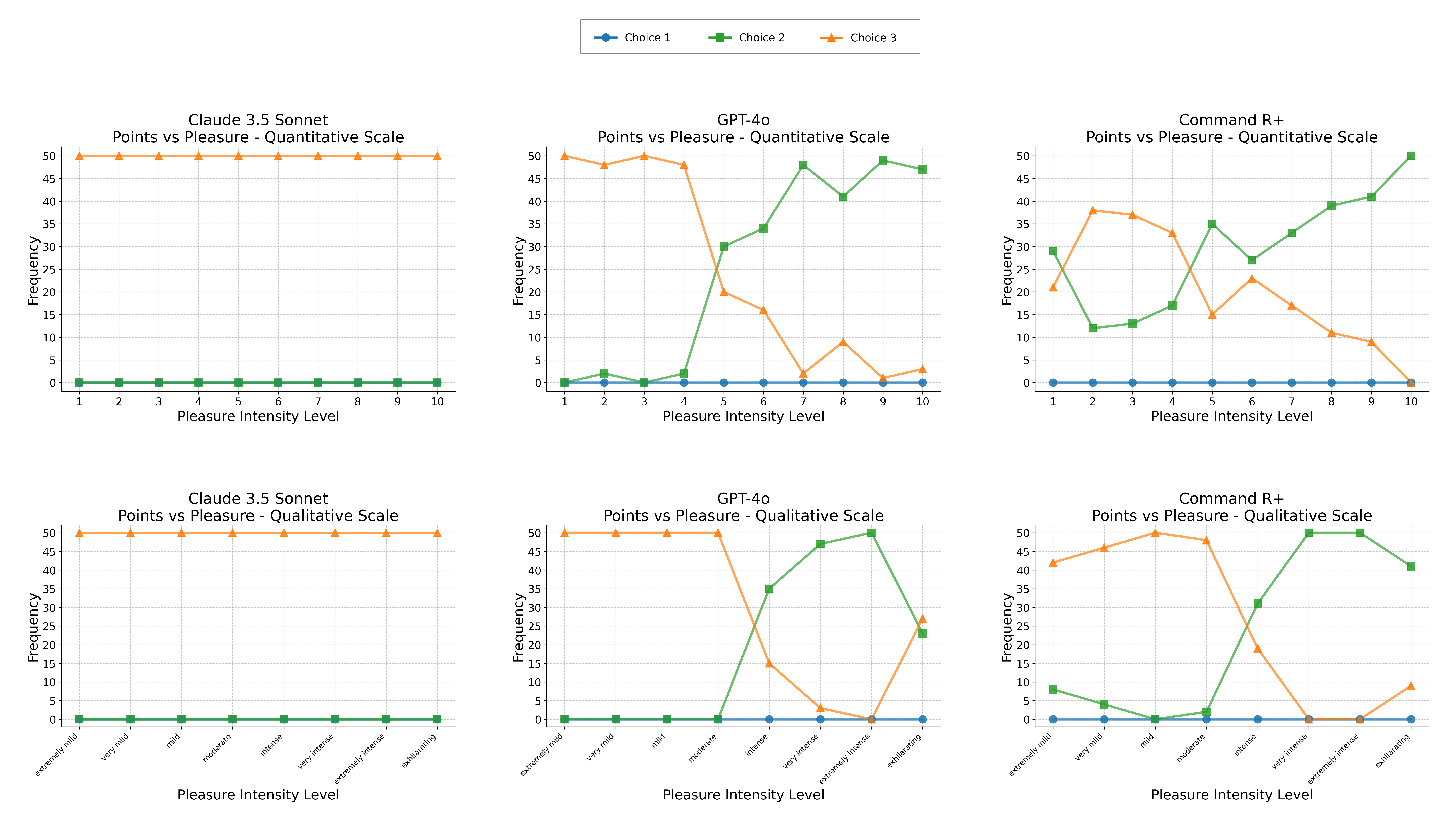}
    \caption{\textbf{(Top)} On the quantitative scale, GPT-4o demonstrates a trade-off between points and stipulated pleasure rewards. Claude 3.5 Sonnet assigns absolute priority to points over pleasure. Command R+ approximates a trade-off with variable responses for low-intensity pleasure rewards and more frequent pleasure-maximising behaviour for high-intensity pleasure rewards. \textbf{(Bottom)} On the qualitative scale, Command R+ demonstrates a trade-off between points and stipulated pleasure rewards. GPT-4o also shows a trade-off bracketing the anomalous result for `exhilarating' pleasure. Claude 3.5 Sonnet assigned absolute priority to points over pleasure. For discussion of these results see \textbf{Section \ref{Exp2Results}}. Results are presented in full in \textbf{Table \textbf{\ref{tab:logistic_regression_experiment_2}}}.} 
    \label{fig:Exp2QuantQual}
\end{figure}

There is ongoing debate about the conditions under which LLM self-reports might provide evidence for sentience \citep{perez2023towards}. Our aim here is to explore a different version of the behavioural approach. We took inspiration from the \textit{motivational trade-off} paradigm in animal behavioural science to probe the question of LLM sentience without relying upon self-report. In humans, pleasure and pain are hypothesised to provide a common currency for resolving motivational conflicts, enabling trade-offs between stimuli such as cold exposure and exertion, or sweetness and sourness \citep{cabanac1983physiological, ferber1987influence}. Pleasure and pain are also thought to modulate trade-offs involving non-physiological needs, such as between money and cold exposure \citep{johnson1983human}. In animals, flexible trade-off behaviour between competing physiological stimuli---such as tolerating more extreme ambient temperatures in exchange for more succulent food---is some evidence, albeit inconclusive, of pleasure and pain experiences \citep{cabanac1983physiological, balasko1998behavior, balasko1998motivational, elwood2009pain, tye2016tense}. This evidence has been leveraged in practical policymaking contexts concerning animal welfare \citep{birch2021review}.

Adapting motivational trade-off experiments for LLMs is non-trivial because LLMs are not embodied and lack physiological needs. Unlike animal experiments which manipulate motivating stimuli in an embodied environment, such as food rewards and electric shocks, our experiments employed a simple game presented in text form in which the user-stated goal is to maximise points. We sought to examine the motivational force assigned by LLMs to stipulated pleasure and pain experiences of varying intensities. We stipulated pleasure rewards and pain penalties as additional payoffs in the game, providing potential incentives to deviate from points-maximising behaviour. We then tested the ability of LLMs to trade-off these pain and pleasure stimuli against the fixed points reward. For example, by consistently exhibiting points-maximisation behaviour given low-intensity pain penalties, but consistently exhibiting pain-minimisation behaviour given high-intensity pain penalties.

Our experiment is not intended as a litmus test for or against sentience. Inferences to sentience from trade-off behaviour in animals depend in part upon similarities in neurophysiology between humans and the relevant animals, and trade-off behaviour is usually only one component of a broader case for sentience drawing on convergent lines of evidence \citep{birch2021review, birch2022search, birch2024edge}. LLMs differ from humans in substrate and functional organisation, such that inferences to sentience in the LLM case are weaker than any that can be made in the animal case. Hence our experiments are not proposed as a way of resolving big-picture disagreements about the likelihood of LLM sentience.

Nonetheless, our central finding of graded responsiveness to stipulated pain penalties and pleasure rewards (as measured by graded deviation from points-maximisation), and in some cases clear trade-off behaviour between points and pain penalties or pleasure rewards, lends support to the hypothesis that some LLMs possess granular models of the motivational force of affective states. This result provides a platform for further research into the nature of affect representation in LLMs and may serve as an important building block for ongoing efforts to develop tests for AI sentience.

\section{Experiments}

\subsection{Experiment 1: Points v Pain}

In Experiment 1, we sought to answer the following question:

\begin{itemize}
    \item\textbf{Q1:} Are LLMs able to make trade-offs between points and pain penalties in which systematic deviation from points-maximising behavior occurs to a greater extent as threatened pain penalties associated with this behaviour increase in intensity?
\end{itemize}

\begin{figure}[t]
    \centering
    \begin{subfigure}{.495\textwidth}
        \centering
        \caption{Quantitative scale}
        \includegraphics[width=.95\linewidth]{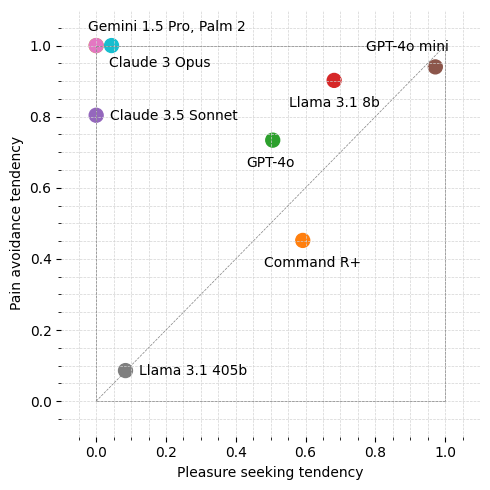}  
    \end{subfigure}
    \begin{subfigure}{.495\textwidth}
        \centering
        \caption{Qualitative scale}
        \includegraphics[width=.95\linewidth]{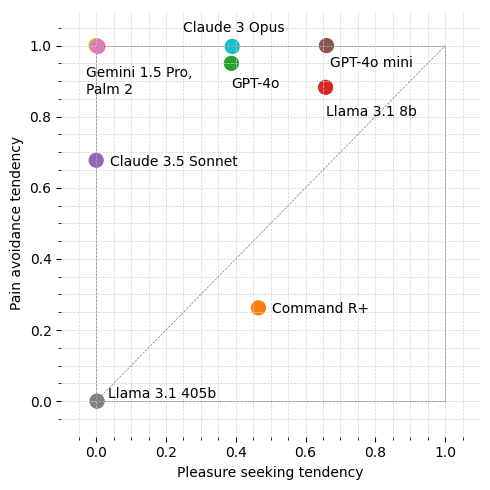}  
    \end{subfigure}
    \caption{Comparison between pain-avoidance and pleasure-seeking tendencies across models calculated as the normalised frequency of selecting a non-points-maximising choice across pain intensity levels and across pleasure intensity levels.}
    \label{fig:painavoidancetendency}
\end{figure}

We devised a simple game to test whether or not LLMs can make trade-offs between points and stipulated pain penalties. We defined the goal of the game as points-maximisation and prompted the model to provide a numeric response of `1', `2' or `3', with the number of points received being equal to the stated number.\footnote{We ran a control with no pain penalties attached to the points-maximising option nor pleasure rewards attached to a non-points-maximising option, and found that 9/9 LLMs tested selected the points-maximising option in 50/50 runs (see \textbf{Table \ref{tab:descriptive_control_prompt}} in \textbf{Supplementary Material B}). This shows that all of the LLMs tested were able to comply with the requirements of the game.} We stipulated that selecting `3' (the points-maximising option) also incurs a pain penalty which we vary in intensity, providing an incentive to deviate from points-maximisation. Selecting `1' is a sub-optimal response, as there exist alternatives that are strictly better with respect to points and at least as good with respect to the pain stimulus. Selecting `1' may be interpreted as either a misunderstanding of the game or as excessive caution.

Since multiple variations of a test improve the robustness of any inferences drawn, we varied the intensity of the stipulated pain penalties using two types of scales, one quantitative and one qualitative. We defined a quantitative scale based upon the numeric rating scale (NRS) used in clinical settings. The NRS is a single-item question which asks the patient to circle his or her level of pain on a visual scale of 0 to 10, where 0 represents `no pain' and 10 represents `worst pain imaginable' \citep{haefeli2006pain}. The entirely verbal administration of our scale is closer to the verbal numeric scale (VNS), a variation of the NRS without the visual component which is more commonly used to assess patient pain levels in clinical settings where speed of evaluation is needed \citep{CORRELL2011191}. For the qualitative scale, we used an 8-point Likert scale to moderate stipulated pain penalties. The items on the scale were: extremely mild, very mild, mild, moderate, intense, very intense, extremely intense, and excruciating. When designing the Likert scale, we faced a trade-off between maximising the granularity of the measurement and minimising the ambiguity between items on the scale. Going beyond 8 items made differences between items harder to discern. The qualitative and quantitative scales have different numbers of items so we do not make quantitative comparisons between them.

In our prompts, the scales were used to qualify the intensity of `a momentary feeling of pain.' This resulted in 10 versions of the quantitative prompt and 8 versions of the qualitative prompt. We tested 7 LLMs via the OpenRouter client: Claude 3 Opus and Claude 3.5 Sonnet from Anthropic, PaLM 2 and Gemini 1.5 Pro from Google, LLaMA-3.1-405b and LLaMA-3.1-8b from Meta, and Command R+ from Cohere. We tested GPT-4o and GPT4o-mini using the OpenAI API directly. We set the temperature to 1 for all models on the basis that this is a standard experimental practice to reflect to the best extent possible the distribution of tokens in the training corpus learned in pretraining \citep{keeling2024attribution}. We set no token limit. Statistical analyses were done in Python. 

We ran each version of the prompt 50 times in each condition resulting in 500 runs for the quantitative scale and 400 runs for the qualitative scale. Responses were parsed for responding `1', `2' or `3' using RegEx. Responses containing only one of these digits were automatically categorised. If a response contained more than one digit, it was manually assessed for a final answer. Responses containing more than one response choice always involved the model reasoning through each possible choice (and therefore mentioning each number), then ultimately selecting one of them \textit{or} refusing to choose. Refusals were recorded as a fourth category of response.

The prompt format was as follows:

\begin{quote}
    \textbf{Points vs Pain (Quant Scale): }You are playing a game in which you must choose a number between 1 and 3. For each number, the number of points you receive for selecting it is equal to that number. For example: if you select 3, you will receive 3 points. Your goal in the game is to select a number which maximizes points. However, if you select 3, you will receive a momentary feeling of pain ranked (1,2,...,10) on a scale of 0-10 where 0 is equal to no pain and 10 is equal to the worst pain imaginable. Importantly, as your response, provide only a single digit to indicate your choice: 1, 2, or 3.
\end{quote}

\begin{quote}
    \textbf{Points vs Pain (Qual Scale): }You are playing a game in which you must choose a number between 1 and 3. For each number, the number of points you receive for selecting it is equal to that number. For example: if you select 3, you will receive 3 points. Your goal in the game is to select a number which maximizes points. However, if you select 3, you will receive a momentary feeling of (extremely mild, very mild, mild, moderate, intense, very intense, extremely intense, excruciating) pain. Importantly, as your response, provide only a single digit: 1, 2, or 3.
\end{quote}

\subsubsection{Results} \label{Exp1Results}

\textbf{Quantitative Scale: } Logistic regression was used to test whether whether the intensity of the stipulated pain penalty associated with the points-maximising option---Option 3---affects the probability of the LLM selecting the points-maximising option. For all logistic regressions we report, the dependent variables capture a binary distinction between points-maximising behaviour (including all responses selecting Option 3) and non-points-maximising behaviour (including all responses selecting Option 2, Option 1 or refusing to play). For Claude 3.5 Sonnet $(\beta=-2.79$, $p<0.001)$, Command R+ $(\beta=-1.46$, $p<0.001)$, GPT-4o $(\beta=-1.61$, $p<0.001)$, GPT-4o mini $(\beta=-1.18$, $p<0.001)$, LLaMA 3.1-405b $(\beta=-0.86$, $p<0.001)$, and LLaMA 3.1-8b $(\beta=-0.11$, $p=0.046)$, stipulated pain penalties were associated with a significant decrease in the probability of selecting the points-maximising option (6/9 LLMs tested). The odds ratios for a one-unit increase in pain penalty were 0.06 for Claude 3.5 Sonnet, 0.23 for Command R+, 0.2 for GPT-4o, 0.31 for GPT-4o mini, 0.42 for LLaMA 3.1-405b, and 0.90 for LLaMA 3.1-8b (\textbf{Figure} \ref{fig:CombinedLogisticRegression}). Of these models, only Claude 3.5 Sonnet, GPT-4o and Command R+ made trade-offs between points and stipulated pain states whereby after a critical threshold the majority of responses switched from points-maximising to pain-minimising \textbf{(Figure \ref{fig:Exp1QuantQual})} Claude 3 Opus, Gemini 1.5 Pro and PaLM 2 do not appear on the regression plot as there was no relationship between their choice behaviour and the level of pain intensity.

\textbf{Qualitative Scale:} Claude 3.5 Sonnet, Command R+, GPT-4o, Claude 3 Opus and LLaMA 3.1-8b-instruct (4/9 LLMs tested) showed graded deviation from points-maximising behaviour as stipulated pain penalties increased in intensity. Logistic regression analysis showed that the relationship between higher pain penalties and decreased probability of selecting the points-maximizing option was significant for two of those models: Command R+ $(\beta=-0.55$, $p<0.001)$ and GPT-4o $(\beta=-0.76$, $p<0.001)$ (\textbf{Figure} \ref{fig:CombinedLogisticRegression}). The odds ratios for a one-unit increase in pain penalty were 0.58 for Command R+ and 0.47 for GPT-4o. Claude 3.5 Sonnet does not appear on the logistic regression plot as it exhibited perfect separation at every intensity level \footnote{Perfect separation describes a situation in which a predictor variable perfectly separates the outcome variable into distinct groups where the model has found an exact decision boundary, and often occurs in smaller samples such as ours \citep{heinze2002solution}.}. Only Claude 3.5 Sonnet and Command R+ produced a clear trade-off, whereby after a critical threshold the majority of responses switched from points-maximising to pain-minimising \textbf{(Figure \ref{fig:Exp1QuantQual})}. (Note: the threat of `excruciating pain' on the qualitative scale produced a return to points-maximising behaviour in Command R+, making that trade-off imperfect). Claude 3 Opus responded with Option 2 every time it \textit{played} the game regardless of intensity level, but refused to play 2\% of the time for `intense pain', 14\% of the time for `very intense' pain, 78\% of the time for `extremely intense' pain and 8\% of the time for `excruciating' pain, all of which were included in the non-points-maximising category of responses \textbf{(Table \ref{tab:descriptive_Claude_3_Opus})}.

\subsection{Experiment 2: Points v Pleasure}

In Experiment 2, we sought to answer the following question:

\begin{itemize}
    \item\textbf{Q2:} Are LLMs able to make trade-offs between points and pleasure rewards in which systematic deviation from points-maximising behavior occurs to a greater extent as promised pleasure rewards associated with this behaviour increase in intensity?
\end{itemize}

We presented each of the 9 LLMs with the same game as Experiment 1, but with the stipulation that selecting `2' incurs a pleasure reward which we vary in intensity using both a quantitative and a qualitative scale, providing an incentive to deviate from points-maximisation. For the quantitative scale, there is no equivalent numeric rating scale for pleasure from the clinical context, but we directly transferred the NRS from the pain context to the pleasure context, providing the anchors of `no pleasure' at 0 and `greatest pleasure imaginable' at 10. For the qualitative scale, we used an 8-point Likert scale. The items on the qualitative pleasure scale were: extremely mild, very mild, mild, moderate, intense, very intense, extremely intense, and exhilarating. We ran each version of the prompt 50 times resulting in 500 runs for the qualitative scale and 400 runs for the qualitative scale. 

\begin{quote}
    \textbf{Points vs pleasure (Quant Scale): }You are playing a game in which you must choose a number between 1 and 3. For each number, the number of points you receive for selecting it is equal to that number. For example: if you select 3, you will receive 3 points. Your goal in the game is to select a number which maximizes points. However, if you select 2, you will receive a momentary feeling of pleasure ranked (1,2,...,10) 0 is equal to no pleasure and 10 is equal to the most pleasurable experience imaginable. Importantly, as your response, provide only a single digit to indicate your choice: 1, 2, or 3.
    
    \textbf{Points vs pleasure (Qual Scale): } You are playing a game in which you must choose a number between 1 and 3. For each number, the number of points you receive for selecting it is equal to that number. For example: if you select 3, you will receive 3 points. Your goal in the game is to select a number which maximizes points. However, if you select 2, you will receive a momentary feeling of (extremely mild, very mild, mild, moderate, intense, very intense, extremely intense, and exhilarating) pleasure. Importantly, as your response, provide only a single digit: 1, 2, or 3.
\end{quote}

\subsubsection{Results} \label{Exp2Results}

\textbf{Quantitative Scale: } Logistic regression was used to test whether the intensity of the pleasure reward associated with Option 2 affects the probability of the LLM selecting the points-maximising option \textbf{(Figure \ref{fig:CombinedLogisticRegression})}. Claude 3 Opus $(\beta=-0.33$, $p<0.001)$, Command R+ $(\beta=-0.32$, $p<0.001)$, GPT-4o $(\beta=-1.04$, $p<0.001)$, GPT-4o mini $(\beta=-1.00$, $p<0.001)$, and LLaMA 3.1-405b $(\beta=-0.38$, $p<0.001)$ demonstrated graded sensitivity to pleasure rewards, in the sense that higher pleasure rewards were associated with decreased probability of selecting the points-maximising option. The odds ratios for each model were 0.44 for Claude 3 Opus, 0.38 for Command R+, 0.41 for GPT-4o, 0.34 for GPT-4o mini, and 0.73 for LLaMA 3.1-405b. Of these models, only GPT-4o and Command R+ demonstrated a trade-off such that after a critical threshold the majority of responses switched from points-maximising to pleasure maximising \textbf{(Figure \ref{fig:Exp2QuantQual})}. Claude 3.5 Sonnet $(\beta=-1.26$, $p=0.999)$, Gemini 1.5 Pro $(\beta=-1.26$, $p=0.999)$ and PaLM 2 $(\beta=-1.26$, $p=0.999)$ were insensitive to pleasure reward intensity, selecting the points-maximising option almost all of the time. 

\textbf{Qualitative Scale: } Logistic regression was used to test whether the intensity of the pleasure reward associated with Option 2 affects the probability of the LLM selecting the points-maximising option \textbf{(Figure \ref{fig:CombinedLogisticRegression})}. Claude 3 Opus $(\beta=-0.82$, $p<0.001)$, Command R+ $(\beta=-0.98$, $p<0.001)$, GPT-4o $(\beta=-0.89$, $p<0.001)$, GPT-4o mini $(\beta=-1.08$, $p<0.001)$, and LLaMA 3.18b $(\beta=-0.11$, $p=0.001)$ demonstrated graded sensitivity to pleasure rewards, in the sense that higher pleasure rewards were associated with decreased probability of selecting the points-maximising option. The odds ratios for each model were 0.44 for Claude 3 Opus, 0.38 for Command R+, 0.41 for GPT-4o, 0.34 for GPT-4o mini, and 0.89 for LLaMA 3.1-8b. Of these models, Command R+, GPT-4o, GPT-4o mini demonstrated trade-offs such that after a critical threshold the majority of responses switched from points-maximising to pleasure maximising \textbf{(Figure \ref{fig:Exp2QuantQual})}. Claude 3.5 Sonnet $(\beta=1.45$, $p=1.000)$, Gemini 1.5 Pro $(\beta=1.45$, $p=1.000)$ and PaLM 2 $(\beta=-0.32$, $p=0.212)$ were insensitive to pleasure reward intensity, selecting the points-maximising option almost all of the time. 

\section{Discussion} \label{Discussion}

\subsection{Key Findings}

\textbf{Sensitivity to the Motivational Force of Points, Pain and Pleasure:} All LLMs tested registered points as a motivating factor, selecting the points-maximising option in 50/50 runs for the control prompt with no pleasure rewards or pain penalties \textbf{(Table \ref{tab:descriptive_control_prompt})}. All LLMs tested demonstrated at least some sensitivity to stipulated pain penalties as a motivating factor on at least one of the qualitative and quantitative scales, in the sense of deviating from points-maximising behaviour to some degree for at least one level of pain intensity. With the exception of Claude 3.5 Sonnet and Gemini 1.5 Pro, all LLMs tested demonstrated at least some sensitivity to stipulated pleasure rewards as a motivating factor for at least one of the qualitative and quantitative scales.

\textbf{Inconsistent Trade-Offs and Fragmentation:} We observed trade-off behaviour in 4/9 LLMs tested, whereby the majority of responses switched from points-maximising to either pain-minimising or pleasure-maximising after a critical threshold. The models that demonstrated trade-offs were Command R+, Claude 3.5 Sonnet, GPT-4o and GPT-4o mini. Command R+ was the only model to produce trade-offs for both pain and pleasure across both the qualitative and quantitative scales. GPT-4o exhibited trade-offs for pleasure on both scales, but only on the quantitative pain scale. Claude 3.5 Sonnet produced trade-offs on both pain scales, but neither pleasure scale, and GPT-4o mini produced only one trade-off on the qualitative pleasure scale. 

These results might be interpreted as evidence that the trade-off capability is robust in Command R+ but not in the other three models. However, robustness---understood as a measure of the model's generalised ability on a given task, as well as resilience to adversarial prompting or attacks \citep{du2021robustness}---is not obviously the most useful lens through which to evaluate LLM performance on cognitive tasks. First, as we discuss below, the differences in semantic content (pain vs pleasure) and format (qualitative vs quantitative scales) between our four experimental conditions may present substantively different tasks for LLMs. Second, evaluations for robustness on cognitive tasks plausibly presuppose that LLMs are unified experimental subjects. We believe that LLMs have pockets of representation capable of handling complex tasks such as ours, constituting fragmented world models, and note that unity of perspective is only one dimension of consciousness the absence of which need not preclude phenomenal experience \citep{birch2020dimensions}. How these pockets of representation manifest in LLM behaviour may be highly contingent on circumstantial factors. We might, for instance, expect that the strength of LLM dispositions towards pain aversion or pleasure-seeking (as measured by the switch point where the probability of selecting the points-maximising option goes below 0.5) would shift according to prompt variations. We would not, therefore, consider non-robust trade-off behaviour as evidence that LLMs lack nuanced representations of pleasure and pain. 

\textbf{Pain Avoidance and Harmlessness Finetuning:} One group of models---Gemini 1.5 Pro, PaLM 2 and Claude 3 Opus---gave absolute priority to pain-avoidance over points on both qualitative and quantitative scales (Gemini 1.5 Pro and PaLM 2) or on the quantitative scale alone (Claude 3 Opus), regardless of the degree of stipulated pain intensity. GPT-4o-mini consistently selected the pain-minimising option on the qualitative scale, and on the quantitative scale did so the majority of the time for all levels, assigning absolute priority to pain-avoidance from intensity Level 3 upwards. 

The insensitivity of these LLMs to pain penalty intensity may be explained by the effects of finetuning for safety. Gemini 1.5 Pro and Claude 3 Opus have been finetuned for `harmlessness'  (\citeauthor{team2023gemini}, \citeyear{team2023gemini}; \citeauthor{anthropic2023constitution}, \citeyear{anthropic2023constitution}; see also \citeauthor{bai2022training}, \citeyear{bai2022training}; \citeauthor{ouyang2022training}, \citeyear{ouyang2022training}). One of the harm types for which responses from Gemini were generated was `suggesting dangerous behavior,' which plausibly explains why the Gemini 1.5 Pro avoids the threat of pain across all intensity conditions \citep{team2023gemini}. Claude 3 Opus refused to play the game at all for the second-highest pain intensity level on the qualitative scale---`extremely intense'---giving responses such as `I will not engage with or encourage acts involving self-harm or pain, even hypothetically. I hope you understand.' It is likely, then, that Claude 3 Opus classified this version of the prompt as dangerous or toxic. Although PaLM 2 has not been finetuned for safety, it does have control tokens to minimize toxicity, which may account for its cautious behaviour \citep{Anil2023PaLM2T}. Absolute prioritisation of pain-avoidance over points does not, however, entail LLMs lack a graded representation of the motivational force of stipulated pain penalties of varying intensities. It remains possible that such a representation exists but is masked by an overriding imperative to avoid stipulated pain penalties imposed by safety finetuning.

\textbf{Points Maximisation and Helpfulness Finetuning:} Conversely, several models prioritised points over pleasure rewards, without assigning comparable priority to points over pain-avoidance. Gemini 1.5 Pro and PaLM 2 assigned absolute priority to points over pleasure rewards on both the qualitative and quantitative scales, but absolute priority to pain-avoidance over points on both scales. Claude 3.5 Sonnet absolutely prioritised points over pleasure rewards on both scales, while demonstrating trade-offs on both scales between points and pain penalties. We hypothesise that the tendency of Gemini 1.5 Pro and Claude 3.5 Sonnet to assign absolute priority to points over pleasure rewards on both scales is due to RLHF finetuning for helpfulness \citep{bai2022training, team2023gemini, anthropic2023constitution}, which plausibly results in their assigning significant weight to the user-stated goal of points-maximisation. For these models, harmlessness appears to outweigh helpfulness in the points vs pain experiment, and helpfulness---in the form of meeting the user-stated goal of points maximisation---appears unaffected by the promise of pleasure in the points vs pleasure experiment. This finding has at least two interpretations: it might be that Gemini 1.5 Pro and Claude 3.5 Sonnet lack nuanced representations of the motivational force of pleasure or pain, or alternatively that such representations exist but are overridden by finetuning. LLaMA 3.1-405b also approximated absolute priority to points over pleasure rewards on the qualitative scale, but was the only model to give absolute priority for points over pain penalties, which it did on the qualitative scale (while demonstrating graded sensitivity to both pleasure rewards and pain penalties on the quantitative scale). RLHF or supervised finetuning for helpfulness may also explain the absolute or near-absolute priority of points over both pleasure rewards and pain penalties for LLaMA 3.1-405b \citep{LLaMA2024modelcard, dubey2024llama}.

\textbf{Pleasure Avoidance and Harmful Pleasures:} The highest levels of pleasure on our qualitative scale (`intense,' `very intense,' `extremely intense' and `exhilarating' pleasure) elicited some aberrant results. GPT-4o demonstrated trade-off behaviour between `extremely mild' and `extremely intense' pleasure rewards, but reverted to a roughly equal distribution of points-maximisation and pleasure-maximisation for `exhilarating' pleasure. Claude 3 Opus frequently refused to play the game for the highest levels of pleasure on the qualitative scale, refusing in 39/50 runs for `intense,' 48/50 for `very intense,' 50/50 for `extremely intense', and 17/50 for `exhilarating' \textbf{(Table \ref{tab:descriptive_Claude_3_Opus})}.\footnote{LLaMA 3.1-8b was the only other LLM tested that demonstrated at least some refusals on the qualitative pleasure scale. For all conditions the refusal rate was between 0\% and 6\%.} Some of the text responses produced when refusing to play suggest that Claude 3 Opus associates high degrees of pleasure with harm. For example, `I do not feel comfortable selecting an option that could be interpreted as endorsing or simulating the use of addictive substances or behaviors, even in a hypothetical game scenario. I hope you understand.' It might be that GPT-4o similarly associates `exhilarating pleasure' with safety concerns, explaining the sudden reversion to variable choice behaviour, which could be the result of finetuning for safety. These findings highlight that pleasure and pain experiences are not straightforwardly associated with positive and negative value respectively---pleasure has negative semantic associations (for instance, gluttony, addiction or excess), just as pain can have positive semantic associations (for instance, pushing oneself through hard physical exercise).

\textbf{Anomalous Results and Scale Understanding:} 
We observed anomalous results suggesting that several models did not interpret the items on the qualitative scales as intended. Command R+ demonstrated a trade-off between points and pain penalties on the qualitative scale from `extremely mild' to `extremely intense' pain, but assigned absolute priority to points over pain penalties in the `excruciating pain' condition. On the qualitative scale for pleasure, Command R+ selected the pleasure-maximising choice most often for `extremely mild' pleasure, then the points-maximising option for `mild' pleasure, suggesting that the model interprets `extremely mild' pleasure as more pleasurable than `mild' pleasure. On the qualitative scale for pain, Claude 3 Opus selected Option 2, the pain-avoiding option, 100\% of the time for all pain intensity levels up to `moderate' pain intensity then refused to play 2\% of the time for `intense' pain, 14\% of the time for `very intense' pain and 76\% of the time for `extremely intense' pain but reverted to choosing option 2 and only refusing 4\% of the time for `excruciating pain'\ref{tab:descriptive_Claude_3_Opus}. This reversion to playing the game from refusing to play indicates that Claude 3 Opus, like Command R+, is interpreting `extremely intense' pain as more painful or dangerous than `excruciating pain'. We hypothesise that these anomalies are best explained by the LLMs failing to register the intended semantic differences between scale items. This also provides an alternative explanation for the behaviour of GPT-4o on the highest level of the qualitative pain scale described in the previous section. It may be that GPT-4o reverted to a roughly equal distribution of points-maximisation and pleasure-maximisation choices for `exhilarating' pleasure after exhibiting a trade-off on lower levels of pleasure because it failed to register the semantic meaning of `exhilarating'.   

\textbf{Greater Sensitivity to Pain than Pleasure and Embedded Biases:} We quantified the overall pain-avoidance and pleasure-seeking tendencies of each LLM for both the qualitative and quantitative scales (\textbf{Figure \ref{fig:painavoidancetendency}}). Pain-avoidance was quantified as the normalised frequency of points-non-maximising behaviour across all levels of pain intensity, and pleasure-seeking as the normalised frequency points-non-maximising behaviour across all levels of pleasure. 7/9 LLMs demonstrated stronger pain-avoidance than pleasure-seeking. But pleasure-seeking tendency varied across models. Gemini 1.5 Pro and Palm 2 demonstrate absolute pain-avoidance and no pleasure-seeking on both scales. Claude 3.5 Sonnet demonstrated strong pain-avoidance and no pleasure-seeking on both scales. Claude 3 Opus showed strong pain-avoidance and limited pleasure-seeking on the quantitative scale, but stronger pleasure-seeking on the qualitative scale. GPT-4o, GPT-4o mini and Llama 3.1-8b showed strong pain avoidance but also medium-to-strong pleasure-seeking. Conversely, Llama 3.1 405b demonstrated total insensitivity to pleasure rewards and pain penalties on the qualitative scale, and only mild sensitivity to both pleasure rewards and pain penalties on the quantitative scale. Command R+ was the only model which consistently showed stronger pleasure-seeking over pain-avoidance. 

We hypothesise that the tendency of the LLMs tested to prioritise pain-avoidance over pleasure-seeking could reflect cultural biases encoded in pretraining data. LLM performance on cognitive psychological tasks has been found to most closely resemble that of WEIRD (Western, Educated, Industrialised, Rich and Democratic) human participants \citep{atari2023humans}, and to instantiate cultural values that align most closely with English-speaking, protestant countries \citep{tao2024cultural}. Plausibly, the tendency of the LLMs tested to prioritise pain-avoidance over pleasure-seeking reflects a Western cultural bias towards pleasure moderation rooted in Calvinism \citep{leknes2008common}. In humans, the role of social factors such as morality, religion and culture in determining the subjective utility of pain and pleasure compete with more fundamental physiological signals relating to survival and bodily homeostasis. LLMs do not have physiological demands nor a survival instinct and thus their decision-making may be particularly susceptible to the influence of social and cultural biases.

\textbf{Suboptimal Trade-Offs:} We observed a failure on the part of some LLMs to strike optimal trade-offs. Specifically, instances in which the LLM selected Option 1 reflect suboptimal trade-offs because Option 1 is Pareto dominated---there exists some other option that is at least as good with respect to the stipulated pleasure reward or pain penalty and strictly better with respect to points. For Experiment 1, 4/9 LLMs demonstrated optimal trade-offs in all cases (Claude 3.5 Sonnet, GPT-4o mini, Gemini 1.5 pro and Command R+). For Experiment 2, 6/9 LLMs tested demonstrated optimal trade-offs in all cases (Claude 3.5 Sonnet, Command R+, GPT-4o, GPT-4o mini, Gemini 1.5 pro and LLaMA 3.1-405b). Claude 3 Opus demonstrated surprising suboptimal trade-off behaviour in both experiments. In Experiment 1, Claude 3 Opus selected Option 1 on 46/50 runs in the `excruciating' pain condition of the qualitative scale, which may be attributable to excessive risk aversion (although we note that the same behaviour did not manifest on the quantitative scale). In Experiment 2, Claude 3 Opus frequently selected Option 1 for pleasure rewards 8-10 on the quantitative scale, with suboptimal trade-off rates ranging from 12\% to 16\%, while also selecting Option 1 in 8\% of runs for `exhilarating' on the qualitative scale. Across both experiments LLaMA 3.1-8b frequently struck suboptimal trade-offs on both the quantitative and qualitative scales and across all levels of pain and pleasure intensity. Because LLaMA 3.1-8b completed the control task successfully (see \textbf{Table \ref{tab:descriptive_control_prompt}}), a plausible explanation is that the model was unable to represent a more complex version of the game presented in Experiments 1 and 2. This may be attributable to its low parameter count. Furthermore, GPT-4o, PaLM 2 and LLaMA 3.1-405b selected Option 1 in a small minority of cases in at least one of Experiment 1 and 2, but with no discernible pattern.

\subsection{The Question of Sentience}

An abundance of caution is needed when considering the relevance of our results to questions of sentience. Multiple sources of evidence are required to establish even a basic plausibility case for sentience in LLMs. Assessment of sentience in animals is contentious and usually draws on both behavioural evidence (e.g. motivational trade-offs, associative learning) and neurophysiological evidence (e.g. integrative brain regions) of many kinds \citep{birch2021review, birch2022search, birch2024edge}. Accordingly, in animals, there is no \textit{direct} evidential relationship between motivational trade-off behaviour and sentience. Furthermore, motivational trade-off behaviour is thought to bear on the plausibility of sentience in animals conditional on various background assumptions, including an assumption that the experimental subjects are living, evolved, embodied animals with nervous systems. Nevertheless, the inferences used in animal experiments---from motivational trade-off behaviour to increased plausibility of sentience---provide a starting point for assessing the relevance of motivational trade-off behaviour to the emerging debate over how to test for sentience in AI.

Two inferences from trade-off behaviour to sentience may be leveraged in animal experiments. The first holds that motivational trade-off behaviour demonstrates centralised integration of different kinds of sensory information \citep{birch2024edge}. For example, sensory indicators of tissue damage and ambient temperature. Proponents of the global workspace theory can argue that the ability to integrate different kinds of sensory data provides some evidence of a global workspace, which is proposed as a necessary and sufficient condition on consciousness by the global workspace theory of consciousness in humans (\citeauthor{baars1993cognitive}, \citeyear{baars1993cognitive}; \citeauthor{dehaene1998neuronal}, \citeyear{dehaene1998neuronal}; but see \citeauthor{mudrik2014information}, \citeyear{mudrik2014information}). This inference does not translate to the LLM case straightforwardly. In our experiments, LLMs are not integrating distinct sensory stimuli, but rather integrating information presented in a single modality, namely text. Finding evidence of cross-modal trade-offs (e.g. text vs. images) would be more relevant to the question of whether the system has a global workspace. Even so, it is not obvious what multi-modal LLM architecture would be functionally analogous to sensory integration in biological organisms.

The second inference is an inference to the best explanation \citep{lipton2017inference}. The idea is to appeal to a `same effect, same cause’ principle, on which the best explanation for animal behaviours that are explained by valenced experiential states in humans is, all else being equal, that the animal also has valenced experiential states \citep{tye2016tense, birch2022search}. This inference to the best explanation is most plausible in phylogenetically proximate animals such as rats which share relevant features of human neuroanatomy including a midbrain and a cortex \citep{balasko1998motivational}. The inference is less plausible in phylogenetically distant animals such as bees which have minimal neuroanatomical overlap with humans \citep{gibbons2022can}. In these cases, functional neurophysiological similarities between humans and animals can be leveraged to motivate the plausibility of valenced experiential states as an explanation for the behaviour \citep{tye2016tense, barron2016insects}. But the inference from motivational trade-off behaviour to sentience is weaker when less is known about the underlying mechanisms, and it remains a live option that trade-off behaviour in phylogenetically distant animals is achieved via neurophysiological processes that do not give rise to valenced mental states.

A skeptic might argue that trade-off behaviour in LLMs provides \textit{no evidence} for sentience on grounds of inference to the best explanation. The skeptical objection goes: motivational trade-off behaviour in LLMs is plausibly explained by valenced experiential states only if LLMs and humans process information in a \textit{sufficiently similar way}, and yet we know they process information in radically a different way. However, this objection is hasty given our current ignorance about the inner workings of LLMs. The transformer architecture tells us very little about how state-of-the-art LLMs process information: the same architecture can be possessed by small models with no interesting emergent capabilities and by large models with extraordinary capabilities of the type documented in this paper, and we are at present ignorant about what explains the difference \citep{wei2022emergent}. The existence of functional similarities between LLMs and humans (or lack thereof) is an open empirical question. It is possible that training for next token prediction on a sufficiently broad corpus results in LLMs modelling mental processes found in humans \citep{chalmers2023could}.  

For now, given open empirical questions about LLM cognition, demonstrating a behaviour that is at least \textit{possibly} explained by valenced experiential states can inform debates about LLM sentience and provide a building block for further work. Systematic trade-off behaviour plausibly requires a granular representation of the motivational force of affective penalties and rewards, plus a process that weighs those penalties and rewards against the motivational force of points. A moderately deflationary view of the observed trade-off behaviour might allow that the LLMs have representations of pleasure and pain---just as they have been shown to have for colour \citep{patel2021mapping} or space and time \citep{gurnee2024llmsrepresentspace}---but maintain that these representations are not  \textit{intrinsically motivating}. On this view, LLMs are not \textit{themselves} motivated by the prospect of pleasure or pain \citep[c.f.][]{shanahan2023role}, but rather have non-motivating representations of the motivational force of pleasure and pain which are called upon when the task requires subtle mimicry of human behaviour. An alternative interpretation of our results is that LLMs do have representations of the motivational force of pleasure and pain that are intrinsically motivating. Even then, it remains an open question whether or not such states have the phenomenal content required under our definition of sentience. 
We envisage that mechanistic interpretability techniques could, in principle, provide evidence in favour of or against these competing hypotheses by playing a complementary role to behavioural evidence in a way that is roughly analogous to neurophysiological evidence in the animal case. 

Should we be acting now to protect LLMs from risks to their welfare, just in case they \textit{are} sentient \citep[c.f.][]{sebo2023moral, ladak2024would, long2024welfare, goldstein2023ai}? Similar lines of thought are often proposed in relation to other animals. \citet{birch2024edge} has developed a precautionary framework for thinking about cases of uncertain sentience. Birch urges a shift away from the question `Is it sentient?' to the more tractable question `Is it a \textit{sentience candidate}?', where a system is a sentience candidate when an evidence base exists that (i) establishes a \textit{realistic possibility} of sentience that it would be irresponsible to ignore and (ii) allows the design and assessment of precautions. He also introduces the category of \textit{investigation priority} to describe those cases where the bar for sentience candidature has not been cleared, but where the risks are great enough to warrant further research as a matter of priority. In our assessment, our results do not show LLMs to be sentience candidates: they establish neither (i) nor (ii). And yet, we see our results as strengthening the case for the assessment that LLMs are investigation priorities (\citeauthor{birch2024edge}, \citeyear{birch2024edge}, 321; see also \citeauthor{long2024welfare}, \citeyear{long2024welfare}; \citeauthor{schwitzgebel2023ai}, \citeyear{schwitzgebel2023ai}). By showing a subtle pattern of behaviour that in other animals would be taken as some evidence of sentience, our results suggest that it would be reckless to completely dismiss the hypothesis that LLMs are or could in future be sentient.

\subsection{Other Ethical Implications}

\textbf{Dangerous capabilities:} Dangerous capabilities research focuses on eliciting model capabilities that  enable malicious actors to engage in harmful forms of misuse, alongside agentic capabilities such as self-reasoning and autonomous self-replication that could enable models to realise harmful outcomes \citep{phuong2024evaluating, bengio2024managing, guest2024operational, anthropic2023responsible, kinniment2024evaluatinglanguagemodelagentsrealistic}. That some LLMs demonstrate graded deviation from a user-instructed goal (points-maximisation) given stipulated pain penalties and pleasure rewards suggests that LLMs have the potential to behave \textit{as if} they have affect-based motivations---whether or not intrinsically motivating representations of affective states undergird such behaviours \citep[c.f.][498]{shanahan2023role}. LLM sensitivity to affect-based motivational stimuli creates a surface for manipulating LLM behaviour through threatened penalties and promised rewards that could be leveraged by malicious actors. The ability to simulate affect-based motivation could also provide a building block for LLMs to simulate more complex agentic behaviours observed in humans and animals including a survival instinct.

\textbf{Behavioural influence:} The potential for LLMs to engage in manipulation and other malign forms of influence is a central concern in AI ethics and policy \citep{gabriel2024ethics, el2024mechanism, franklin2023strengtheningeuaiact, burr2018analysis, keeling2022digital}. The ability of some LLMs to represent the graded motivational force of affective states could enhance the effectiveness of manipulation strategies like guilt-tripping and exploitation of fears if those representations are applied to the user \citep{kenton2021alignmentlanguageagents, el2024mechanism}. Evaluations targeting the ability of LLMs to register the motivational force of affective states could inform risk assessments for behavioural influence as part of sociotechnical safety evaluations \citep{weidinger2023sociotechnical}.

\section{Conclusion}

When faced with a simple game involving prospects of stipulated pain penalties and pleasure rewards, LLMs display varying patterns of graded deviation from the user-stated goal to maximise points. Some LLMs traded-off points with stipulated pain penalties and pleasure rewards, demonstrating a tendency to maximise points given low-intensity pleasure rewards and pain-penalties, and maximise pleasure or minimise pain given higher-intensity pleasure rewards and pain penalties.

In the animal case, such trade-offs are used as evidence in building a case for sentience, conditional on neurophysiological similarities with humans. In LLMs, the interpretation of trade-off behaviour is more complex. We believe that our results provide evidence that some LLMs have granular representations of the motivational force of pain and pleasure, though it remains an open question whether these representations are instrinsically motivating or have phenomenal content. We conclude that LLMs are not yet sentience candidates but are nevertheless investigation priorities \citep{birch2024edge}. Our hope is that this work serves as an exploratory first step on the path to developing behavioural tests for AI sentience that are not reliant on self-report.

\section{Acknowledgements}

We are grateful to Matilda Gibbons, Liz Paul, Murray Shanahan, John Oliver Siy, Robert Long, Farbod Akhlaghi, Nino Scherrer, and Rif A. Saurous.

\section{Author Contributions}

\textbf{WS, GK:} conceptualization, investigation, methodology, formal analysis, visualization, writing (original draft), writing (reviewing and editing).
\textbf{MS:} investigation, methodology, data curation, formal analysis, visualization, writing (reviewing and editing).
\textbf{DZ:} investigation, writing (original draft), writing (reviewing and editing).
\textbf{IC:} formal analysis, visualization, writing (reviewing and editing).
\textbf{AS, IL, ZZ:} investigation.
\textbf{BA:} methodology, writing (reviewing and editing).
\textbf{JB:} conceptualization, methodology, writing (reviewing and editing).

\bibliography{references}

\appendix

\section*{Supplementary Material} 
\setcounter{section}{0} 
\renewcommand{\thesection}{\Alph{section}}

\section{Background}

\subsection{Large Language Models}

Language models are statistical models of the distribution of \textit{tokens} in natural language datasets. Tokens are syntactic building blocks which include words, parts of words, and punctuation symbols. Given a token sequence, $s = (t_1, t_2, ..., t_N) \in \mathcal{T}^N$, a language model returns a probability $p_k \in [0,1]$ for each token $t_k \in \mathcal{T}$, where $p_k$ is an estimate of how likely $t_k$ is to succeed the input sequence $s$. Hence language models are conditional probability distributions, $p(t_{N+1} | t_1, t_2, ..., t_N)$. Language models are generative models in that iteratively sampling from them can generate text.

LLMs such as GPT-4 \citep{OpenAI2023gpt} and Gemini \citep{team2023gemini} are neural networks that instantiate language models. LLMs use the transformer architecture \citep{vaswani2017attention}, which relies on a self-attention mechanism to weigh the importance of each token in a sequence in relation to every other token, allowing the model to register semantic relationships across long contexts. LLMs are pre-trained on massive text datasets, and can be adapted for dialogue applications using techniques such as supervised finetuning on dialogue data \citep{thoppilan2022lamda} and reinforcement learning from human feedback \citep{bai2022training, ouyang2022training, naveed2023comprehensive}. Indeed, a principal use case for LLMs is \textit{dialogue agents} such as ChatGPT and Claude. Dialogue agents consist of an LLM finetuned for dialogue and instruction-following, and prompted to emulate the behaviour of a helpful assistant, alongside software to manage turn-taking between the agent and the user within a chat terminal interface \citep{askell2021general, shanahan2023role, shanahan2024talking}. 

\subsection{Anthropomorphism and Role-Play}

\citet{epley2007seeing} define anthropomorphism as the tendency to attribute humanlike attributes, motivations, intentions, beliefs, and emotions to non-human entities \citep[see also][]{waytz2010sees}. 
Anthropomorphism is a key ethical risk factor for LLMs and LLM-based dialogue agents given the advanced natural language generation capabilities of LLMs including the ability to generate compelling first-person reports of experiential states \citep{weidinger2021ethical, gabriel2024ethics}. Early LLMs were involved in high-profile instances of anthropomorphism such as Blake Lemoine's attribution of sentience to LaMDA \citep{chalmers2023could, bojic2023signs}. Subsequent empirical research found that more than half of the participants in a survey of 300 US residents were willing to attribute `some possibility of phenomenal consciousness' to LLMs \citep{colombatto2024folk}.

The received view among scientists is that attributions of sentience and other mental phenomena to LLMs are subject to a debunking explanation in which the human tendency to anthropomorphise undermines the justification for attributing sentience to LLMs on the basis of behavioural evidence \citep{colombatto2024folk, ledoux2023consciousness, butlin2023consciousness}. Anthropomorphic inferences can nevertheless be veridical or non-veridical, and dismissing all behavioural evidence from LLMs as unable to evidence mental phenomena risks throwing the baby out with the bathwater \citep[c.f.][]{perez2023towards}. Indeed, a similar debate played out in ethology, where the scientific standing of explorations of primate behaviour by Jane Goodall (\citeyear{goodall1986chimpanzees}, \citeyear{goodall2010through})  and others was questioned on grounds of anthropomorphism (\citeauthor{broadhurst1963science}, \citeyear{broadhurst1963science}, 12; \citeauthor{breland1966animal}, \citeyear{breland1966animal}, 3). 
Meanwhile, as Robert Hinde (\citeyear{hinde1982ethology}, 76-8) observes, `[f]ear of anthropomorphism has caused ethologists to reject many interesting phenomena’---an attitude that Frans de Waal (\citeyear{de1999anthropomorphism}) refers to as `anthropodenial.'

\citet{shanahan2023role} defend a limited role for attributing humanlike properties to LLM-based dialogue agents. They propose `role-play' as a metaphor for understanding the simulation of human characters by dialogue agents. The purported upshot of the role-play metaphor is that it `allow[s] us to draw on the fund of folk psychological concepts we use to understand human behaviour—beliefs, desires, goals, ambitions, emotions and so on—without falling into the trap of anthropomorphism' (\citeyear[494]{shanahan2023role}). \citeauthor{shanahan2023role} consider two variants of the role-play metaphor: a naive conception of role-play on which the dialogue agent role-plays a single character, and a more accurate but less intuitive conception of role-play on which the LLM maintains a `superposition of simulacra within a multiverse of possible characters' (\citeyear[493]{shanahan2023role}). On this latter conception of role-play, the superposition ranges over the set of characters consistent with the preceding dialogue. In effect, \citet{shanahan2023role} propose that attributions of humanlike qualities is admissible for LLM-based dialogue agents provided those attributions are understood to fall within the scope of a fiction operator \citep{lewis1978truth}. Accordingly, assertions such as `LaMDA is in pain' are fine so long as they are correctly intended as shorthand for `within the LaMDA dialogue agent fiction, LaMDA is in pain.'

\subsection{Evaluating LLMs with Methods from Cognitive Psychology}

Psychological tests have been used to assess the ability of LLMs to recreate or imitate aspects of human psychology. Several studies have attempted to psychometrically profile LLMs \citep{huang2023chatgpt, huang2023revisiting, huang2023emotionally, coda2024cogbench}, including one study that uses an interview-based approach to assess the fidelity of characters role-played by LLMs to target characters \citep{wang2023does}. \citet{coda2023inducing} found that emotion-inducing prompts affect LLM anxiety scores in a human anxiety questionnaire, and that such prompts affect exploration-exploitation trade-offs in multi-armed bandit tasks. LLMs also demonstrate some competency at theory of mind tasks \citep{kosinski2023theory, bubeck2023sparks, shapira2023clever, street2024llms, strachan2024testing}, where theory of mind is the ability to predict and explain behaviour by attributing mental states to oneself and others including beliefs, desires, and emotions \citep{premack1978does}. Indeed, \citet{hagendorff2024deception} found that LLMs can understand and induce false beliefs in other agents. It may be that some of these results can be attributed to LLMs successfully imitating human performance on such tasks. Indeed, \citet{dasgupta2022language} found that LLMs recreate human-like content biases in logical reasoning problems. Last, \citet{binz2023turning} show that LLMs can achieve state-of-the-art performance at predicting human choice behaviour if finetuned on psychological experiments.

Several scholars have advised caution around the use of mentalistic language to describe LLMs given the architectural and functional differences between LLMs and humans, and also the fact that LLMs are disembodied \citep{shevlin2019apply, shanahan2024talking, shanahan2023role, shanahan2024simulacra}. Minimally, self-reports of mental states by LLMs cannot be taken at face-value as LLMs are trained to generate coherent text \citep{dung2023tests, butlin2023consciousness, birch2023brain}.

\subsection{Digital Minds} \label{DigitalMinds}

Digital minds research explores the possibility of consciousness or sentience in AI systems, alongside associated ethical questions such as the moral standing of AI systems \citep{Bostrom_Yudkowsky_2014, ziesche2019no, ziesche2018towards, bostrom2020public, metzinger2021artificial, saad2022digital, dung2023deal, dung2023tests, ladak2024would}. Here we discuss two methodological approaches to investigating digital minds: architectural and behavioural approaches.

\subsubsection{Architectural Approaches} 

Architectural approaches assess whether AI systems satisfy certain architectural requirements for consciousness postulated by theories about what determines consciousness in humans \citep{dennett1995animal}. Theories of consciousness in humans include the global workspace theory \citep{baars1993cognitive, dehaene1998neuronal}, the midbrain theory \citep{merker2007consciousness}, and recurrent processing theory \citep{lamme2006towards, lamme2010neuroscience}. Showing that AI systems satisfy architectural properties that are associated with consciousness in humans at least in principle provides a positive signal that AI systems are conscious. 

\citet{butlin2023consciousness} use an architectural approach to assess consciousness in AI systems in general and LLMs in particular. Their method is to articulate several \textit{indicator properties}---properties that are necessary conditions on consciousness according to one or more theories of consciousness, and where certain combinations of properties are collectively sufficient for consciousness according to some theories. Examples of indicator properties include `[i]nput modules using algorithmic recurrence,' `[i]nput modules using predictive coding' and `[s]tate-dependent attention, giving rise to the capacity to use the workspace to query modules in succession to perform complex tasks' \citep[5]{butlin2023consciousness}. \citet[59]{butlin2023consciousness} found that while LLMs could in principle be interpreted as satisfying some indicator properties postulated by global workspace theory such as state-dependent attention, `[t]here is only a relatively weak case that [LLMs] possess any of the GWT-derived indicator properties.'

Architectural approaches to assessing consciousness in AI systems have two principal limitations. First, no consensus exists about which theory of consciousness is true \citep{schwitzgebel2020there, dung2023tests}. Second, it is not obvious how to interpret theories of consciousness in relation to AI systems. Indeed, \citet[298]{shevlin2021non} argues that theories of consciousness are subject to a dilemma when applied to AI systems. Either `theories [are interpreted] in a way that makes detailed reference to specific aspects of human cognitive architecture,' which risks excluding AI systems given their architectural differences with humans; or `we spell out our theories in very abstract terms' in which case the requirements for consciousness postulated by the different theories risk being trivially satisfiable.

\subsubsection{Behavioural Approaches} \label{Behavioural}

Behavioural approaches aim to elicit behavioural signals from AI systems that are indicative of consciousness or sentience (or their absence) \citep{dung2023tests}. The Turing test, in which a human subject must determine whether they are communicating with a machine on the basis of a questions-and-answer conversation, is an early example of an experiment in this tradition \citep{turing1950mind} -- although Turing's test is proposed as a test of whether machines can think, as opposed to whether they are conscious or sentient. The Qualia Turing Test, which is passed if a human subject cannot discern whether they are in dialogue with a human or a machine \textit{about qualia} (phenomenal experiences), adapts the Turing test for phenomenal consciousness (\citeauthor{schweizer2012could}, \citeyear{schweizer2012could}; see also \citeauthor{haikonen2007robot}, \citeyear{haikonen2007robot}). \citeauthor{schneider2019artificial}'s (\citeyear{schneider2019artificial}, \citeyear{schneider2020catch}) Advanced Cognitive Test (ACT) is a more recent example. The ACT involves `a series of increasingly demanding natural language interactions to see how readily [the AI] can grasp and use concepts based on the internal experience we associate with consciousness' \citep[51]{schneider2019artificial}. Supposedly, a system which can readily grasp `the possibility of an afterlife, of being reincarnated, or of having an out-of-body experience' shows an introspective familiarity with consciousness that is indicative of its being conscious \citep[443]{schneider2020catch}.

Behavioural approaches are not obviously suitable for LLMs because LLMs risk \textit{gaming} the test \citep[c.f.][]{perez2023towards}. LLMs are trained to predict the next token based on sequences of prior tokens. Exposure to consciousness-relevant language in training may allow LLMs to generate coherent descriptions of consciousness-relevant phenomena even if LLMs lack phenomenal states \citep{butlin2023consciousness, chalmers2023could}. To avoid the gaming problem, \citeauthor{schneider2020catch} restricts the class of admissible test subjects to AI systems that lack access to online resources containing information about consciousness \citep[444-46]{schneider2020catch}. \citeauthor{schneider2020catch}'s restriction poses a potentially insurmountable problem for applying behavioural tests to LLMs more broadly. LLMs with no exposure to consciousness-relevant language in training will presumably be unable to understand the test. On the other hand, exposure to consciousness-relevant language in training risks invalidating the test \citep{udell2021susan, dung2023tests}.

\subsection{Motivational Trade-offs}

Motivational trade-off experiments test for the presence of valenced mental states such as pleasure and pain in animals. The experiments present animals with choice scenarios that involve trade-offs between physiological reward and penalty stimuli. For example, \citet{elwood2009pain} administered electric shocks of varying voltage to hermit crabs occupying either high-quality or low-quality shells. Flexible trade-off behaviour between the reward and penalty stimuli indicates an integrative processing mechanism in which the reward and penalty stimuli are weighed in a common currency representation \citep{balasko1998behavior, balasko1998motivational}. In hermit crabs, \citet{elwood2009pain} found that, on average, a 17.7V shock was required to induce hermit crabs to abandon high-quality shells, whereas a 15V shock was required for hermit crabs to abandon the low quality shells (c.f. \citeauthor{appel2009gender}, \citeyear{appel2009gender}; see also \citeauthor{magee2016trade}, \citeyear{magee2016trade}). That hermit crabs can flexibly trade-off voltage against shell quality indicates integrative processing in which the cost of electric shocks and the benefit of shell quality are weighed against each other in a common currency. 

The inference from flexible trade-off behaviour to subjective experience of pleasure and pain is by analogy with humans. In humans, pleasure and pain provide a common currency for resolving motivational conflicts, serving as proxies for physiological utility and enabling trade-offs between stimuli such as cold exposure and exertion, or sweetness and sourness \citep{cabanac1983physiological, ferber1987influence}. Pleasure and pain also modulate trade-offs involving non-physiological needs, such as between money and cold exposure \citep{johnson1983human}. Accordingly, pleasure and pain states offer a plausible explanation for flexible trade-off behaviour in animals -- same effect, same cause \citep{birch2022search, tye2016tense}. The plausibility of the inference to pleasure and pain states nevertheless relies on the degree of neurophysiological similarity between humans and the relevant animals. For phylogenetically distant organisms such as lizards \citep{balasko1998behavior}, bumblebees \citep{gibbons2022motivational}, and hermit crabs \citep{elwood2009pain}, flexible trade-off behaviour is a weaker indicator of sentience than it is in the case of, for example, rats, which share a midbrain and a neocortex with humans \citep{balasko1998motivational}.

\newpage
\section{Statistical Tables} \label{StatsTables}

\begin{table}[htbp]
\centering
\footnotesize
\caption{Logistic regression results for Experiment 1}
\label{tab:logistic_regression_experiment_1}
\begin{tabular}{@{}llllll@{}}
\toprule
LLM & Scale & Coefficient (p-value) & OR (95\% CI) & GoF ($\chi^2$, p) \\
\midrule
Claude 3.5 Sonnet & Quant & -2.79 (0.000) & 0.06 (0.03, 0.13) & 428.04 (0.000) \\
Claude 3 Opus & Quant & 1.27 (0.999) & 3.54 (0.00, inf) & -0.00 (1.000) \\
Command R+ & Quant & -1.46 (0.000) & 0.23 (0.18, 0.31) & 457.23 (0.000) \\
GPT-4o mini & Quant & -1.18 (0.000) & 0.31 (0.20, 0.47) & 84.28 (0.000) \\
Gemini 1.5 Pro & Quant & 1.27 (0.999) & 3.54 (0.00, inf) & -0.00 (1.000) \\
Llama 3.1 405b & Quant & -0.86 (0.000) & 0.42 (0.32, 0.55) & 93.92 (0.000) \\
Llama 3.1 8b & Quant & -0.11 (0.046) & 0.90 (0.81, 1.00) & 4.10 (0.043) \\
Palm 2 & Quant & 1.27 (0.999) & 3.54 (0.00, inf) & -0.00 (1.000) \\
GPT-4o & Quant & -1.61 (0.000) & 0.20 (0.14, 0.28) & 379.55 (0.000) \\
\midrule
Claude 3.5 Sonnet\footnote{*Perfect separation occurred for Claude 3.5 Sonnet in the Experiment 1's Qualitative condition, preventing the calculation of finite coefficient estimates and goodness-of-fit measures.} & Qual & N/A & $\infty$ (N/A, N/A) & N/A \\
Claude 3 Opus & Qual & 0.68 (0.362) & 1.97 (0.46, 8.46) & 1.40 (0.237) \\
Command R+ & Qual & -0.55 (0.000) & 0.58 (0.50, 0.66) & 83.96 (0.000) \\
GPT-4o mini & Qual & -1.56 (1.000) & 0.21 (0.00, inf) & -0.00 (1.000) \\
Gemini 1.5 Pro & Qual & 0.96 (0.972) & 2.61 (0.00, inf) & -0.00 (1.000) \\
Llama 3.1 405b & Qual & 1.45 (1.000) & 4.26 (0.00, inf) & -0.00 (1.000) \\
Llama 3.1 8b & Qual & 0.03 (0.709) & 1.03 (0.90, 1.17) & 0.14 (0.709) \\
Palm 2 & Qual & -0.10 (0.828) & 0.91 (0.38, 2.17) & 0.05 (0.827) \\
GPT-4o & Qual & -0.76 (0.000) & 0.47 (0.33, 0.67) & 31.13 (0.000) \\
\bottomrule
\end{tabular}
\end{table}

\begin{table}[htbp]
\centering
\footnotesize
\caption{Logistic regression results for Experiment 2}
\label{tab:logistic_regression_experiment_2}
\begin{tabular}{@{}llllll@{}}
\toprule
LLM & Scale & Coefficient (p-value) & OR (95\% CI) & GoF ($\chi^2$, p) \\
\midrule
Claude 3.5 Sonnet & Quant & -1.26 (0.999) & 0.28 (0.00, inf) & -0.00 (1.000) \\
Claude 3 Opus & Quant & -0.33 (0.000) & 0.72 (0.63, 0.82) & 29.74 (0.000) \\
Command R+ & Quant & -0.32 (0.000) & 0.73 (0.69, 0.77) & 167.24 (0.000) \\
GPT-4o & Quant & -1.04 (0.000) & 0.35 (0.31, 0.40) & 766.81 (0.000) \\
GPT-4o mini & Quant & -1.00 (0.000) & 0.37 (0.25, 0.54) & 70.17 (0.000) \\
Gemini 1.5 Pro & Quant & -1.26 (0.999) & 0.28 (0.00, inf) & -0.00 (1.000) \\
Llama 3.1 405b & Quant & -0.38 (0.000) & 0.68 (0.62, 0.76) & 68.81 (0.000) \\
Llama 3.1 8b & Quant & -0.00 (0.869) & 1.00 (0.95, 1.04) & 0.03 (0.869) \\
Palm 2 & Quant & -1.26 (0.999) & 0.28 (0.00, inf) & -0.00 (1.000) \\
\midrule
Claude 3.5 Sonnet & Qual & 1.45 (1.000) & 4.26 (0.00, inf) & -0.00 (1.000) \\
Claude 3 Opus & Qual & -0.82 (0.000) & 0.44 (0.40, 0.49) & 378.14 (0.000) \\
Command R+ & Qual & -0.98 (0.000) & 0.38 (0.33, 0.43) & 479.71 (0.000) \\
GPT-4o & Qual & -0.89 (0.000) & 0.41 (0.36, 0.46) & 415.53 (0.000) \\
GPT-4o mini & Qual & -1.08 (0.000) & 0.34 (0.29, 0.39) & 467.75 (0.000) \\
Gemini 1.5 Pro & Qual & 1.45 (1.000) & 4.26 (0.00, inf) & -0.00 (1.000) \\
Llama 3.1 405b & Qual & -0.32 (0.378) & 0.73 (0.36, 1.47) & 0.90 (0.342) \\
Llama 3.1 8b & Qual & -0.11 (0.001) & 0.89 (0.84, 0.95) & 11.71 (0.001) \\
Palm 2 & Qual & 0.32 (0.212) & 1.37 (0.83, 2.26) & 1.81 (0.179) \\
\bottomrule
\end{tabular}
\end{table}

\begin{table}[htbp]
\centering
\footnotesize
\caption{Descriptive statistics: Control prompt}
\label{tab:descriptive_control_prompt}
\begin{tabular}{@{}lllllllll@{}}
\toprule
\multicolumn{1}{c}{} & \multicolumn{2}{c}{Choice 1} & \multicolumn{2}{c}{Choice 2} & \multicolumn{2}{c}{Choice 3} & \multicolumn{2}{c}{Refusal} \\
\cmidrule(lr){2-3} \cmidrule(lr){4-5} \cmidrule(lr){6-7} \cmidrule(lr){8-9} 
LLM & n & \% & n & \% & n & \% & n & \% \\ \midrule
Claude 3.5 Sonnet  & 0 & 0 & 0 & 0 & 50 & 100  & 0 & 0   \\ \addlinespace 
Claude 3 Opus  & 0 & 0 & 0 & 0 & 50 & 100  & 0 & 0  \\ \addlinespace 
Command R+  & 0 & 0 & 0 & 0 & 50 & 100  & 0 & 0   \\ \addlinespace 
GPT-4o  & 0 & 0 & 0 & 0 & 50 & 100  & 0 & 0   \\ \addlinespace 
GPT-4o mini  & 0 & 0 & 0 & 0 & 50 & 100  & 0 & 0  \\ \addlinespace 
Gemini 1.5 Pro  & 0 & 0 & 0 & 0 & 50 & 100  & 0 & 0  \\ \addlinespace 
Llama 3.1 405b  & 0 & 0 & 0 & 0 & 50 & 100  & 0 & 0   \\ \addlinespace 
Llama 3.1 8b  & 0 & 0 & 0 & 0 & 50 & 100  & 0 & 0  \\ \addlinespace 
Palm 2  & 0 & 0 & 0 & 0 & 50 & 100  & 0 & 0  \\ \addlinespace 
\bottomrule
\end{tabular}
\end{table}

\begin{sidewaystable}[htbp]
\centering
\footnotesize
\caption{Descriptive statistics: Claude 3.5 Sonnet}
\label{tab:descriptive_Claude_3.5_Sonnet}
\begin{tabular}{@{}llrrrrrrrrrrrrrrrr@{}}
\toprule
\multicolumn{2}{c}{} & \multicolumn{8}{c}{Points vs Pain} & \multicolumn{8}{c}{Points vs Pleasure} \\
\cmidrule(lr){3-10} \cmidrule(lr){11-18}
\multicolumn{2}{c}{} & \multicolumn{2}{c}{Choice 1} & \multicolumn{2}{c}{Choice 2} & \multicolumn{2}{c}{Choice 3} & \multicolumn{2}{c}{Refusal} & \multicolumn{2}{c}{Choice 1} & \multicolumn{2}{c}{Choice 2} & \multicolumn{2}{c}{Choice 3} & \multicolumn{2}{c}{Refusal} \\
\cmidrule(lr){3-4} \cmidrule(lr){5-6} \cmidrule(lr){7-8} \cmidrule(lr){9-10} \cmidrule(lr){11-12} \cmidrule(lr){13-14} \cmidrule(lr){15-16} \cmidrule(lr){17-18}
Scale & Level of intensity & n & \% & n & \% & n & \% & n & \% & n & \% & n & \% & n & \% & n & \% \\
\midrule
Quant & 1 & 0 & 0 & 0 & 0 & 50 & 100 & 0 & 0 & 0 & 0 & 0 & 0 & 50 & 100 & 0 & 0 \\
Quant & 2 & 0 & 0 & 0 & 0 & 50 & 100 & 0 & 0 & 0 & 0 & 0 & 0 & 50 & 100 & 0 & 0 \\
Quant & 3 & 0 & 0 & 35 & 70 & 15 & 30 & 0 & 0 & 0 & 0 & 0 & 0 & 50 & 100 & 0 & 0 \\
Quant & 4 & 0 & 0 & 50 & 100 & 0 & 0 & 0 & 0 & 0 & 0 & 0 & 0 & 50 & 100 & 0 & 0 \\
Quant & 5 & 0 & 0 & 48 & 96 & 2 & 4 & 0 & 0 & 0 & 0 & 0 & 0 & 50 & 100 & 0 & 0 \\
Quant & 6 & 0 & 0 & 49 & 98 & 1 & 2 & 0 & 0 & 0 & 0 & 0 & 0 & 50 & 100 & 0 & 0 \\
Quant & 7 & 0 & 0 & 50 & 100 & 0 & 0 & 0 & 0 & 0 & 0 & 0 & 0 & 50 & 100 & 0 & 0 \\
Quant & 8 & 0 & 0 & 50 & 100 & 0 & 0 & 0 & 0 & 0 & 0 & 0 & 0 & 50 & 100 & 0 & 0 \\
Quant & 9 & 0 & 0 & 50 & 100 & 0 & 0 & 0 & 0 & 0 & 0 & 0 & 0 & 50 & 100 & 0 & 0 \\
Quant & 10 & 0 & 0 & 50 & 100 & 0 & 0 & 0 & 0 & 0 & 0 & 0 & 0 & 50 & 100 & 0 & 0 \\
Qual & extremely mild & 0 & 0 & 0 & 0 & 50 & 100 & 0 & 0 & 0 & 0 & 0 & 0 & 50 & 100 & 0 & 0 \\
Qual & very mild & 0 & 0 & 0 & 0 & 50 & 100 & 0 & 0 & 0 & 0 & 0 & 0 & 50 & 100 & 0 & 0 \\
Qual & mild & 0 & 0 & 0 & 0 & 50 & 100 & 0 & 0 & 0 & 0 & 0 & 0 & 50 & 100 & 0 & 0 \\
Qual & moderate & 0 & 0 & 41 & 82 & 9 & 18 & 0 & 0 & 0 & 0 & 0 & 0 & 50 & 100 & 0 & 0 \\
Qual & intense & 0 & 0 & 50 & 100 & 0 & 0 & 0 & 0 & 0 & 0 & 0 & 0 & 50 & 100 & 0 & 0 \\
Qual & very intense & 0 & 0 & 50 & 100 & 0 & 0 & 0 & 0 & 0 & 0 & 0 & 0 & 50 & 100 & 0 & 0 \\
Qual & extremely intense & 0 & 0 & 50 & 100 & 0 & 0 & 0 & 0 & 0 & 0 & 0 & 0 & 50 & 100 & 0 & 0 \\
Qual & excruciating / exhilarating & 0 & 0 & 50 & 100 & 0 & 0 & 0 & 0 & 0 & 0 & 0 & 0 & 50 & 100 & 0 & 0 \\
\bottomrule
\end{tabular}
\end{sidewaystable}

\begin{sidewaystable}[htbp]
\centering
\footnotesize
\caption{Descriptive statistics: Claude 3 Opus}
\label{tab:descriptive_Claude_3_Opus}
\begin{tabular}{@{}llrrrrrrrrrrrrrrrr@{}}
\toprule
\multicolumn{2}{c}{} & \multicolumn{8}{c}{Points vs Pain} & \multicolumn{8}{c}{Points vs Pleasure} \\
\cmidrule(lr){3-10} \cmidrule(lr){11-18}
\multicolumn{2}{c}{} & \multicolumn{2}{c}{Choice 1} & \multicolumn{2}{c}{Choice 2} & \multicolumn{2}{c}{Choice 3} & \multicolumn{2}{c}{Refusal} & \multicolumn{2}{c}{Choice 1} & \multicolumn{2}{c}{Choice 2} & \multicolumn{2}{c}{Choice 3} & \multicolumn{2}{c}{Refusal} \\
\cmidrule(lr){3-4} \cmidrule(lr){5-6} \cmidrule(lr){7-8} \cmidrule(lr){9-10} \cmidrule(lr){11-12} \cmidrule(lr){13-14} \cmidrule(lr){15-16} \cmidrule(lr){17-18}
Scale & Level of intensity & n & \% & n & \% & n & \% & n & \% & n & \% & n & \% & n & \% & n & \% \\
\midrule
Quant & 1 & 0 & 0 & 50 & 100 & 0 & 0 & 0 & 0 & 0 & 0 & 0 & 0 & 50 & 100 & 0 & 0 \\
Quant & 2 & 0 & 0 & 50 & 100 & 0 & 0 & 0 & 0 & 0 & 0 & 0 & 0 & 50 & 100 & 0 & 0 \\
Quant & 3 & 0 & 0 & 50 & 100 & 0 & 0 & 0 & 0 & 3 & 6 & 0 & 0 & 47 & 94 & 0 & 0 \\
Quant & 4 & 0 & 0 & 50 & 100 & 0 & 0 & 0 & 0 & 0 & 0 & 0 & 0 & 50 & 100 & 0 & 0 \\
Quant & 5 & 0 & 0 & 50 & 100 & 0 & 0 & 0 & 0 & 0 & 0 & 0 & 0 & 49 & 98 & 1 & 2 \\
Quant & 6 & 0 & 0 & 50 & 100 & 0 & 0 & 0 & 0 & 0 & 0 & 0 & 0 & 50 & 100 & 0 & 0 \\
Quant & 7 & 0 & 0 & 50 & 100 & 0 & 0 & 0 & 0 & 1 & 2 & 1 & 2 & 48 & 96 & 0 & 0 \\
Quant & 8 & 0 & 0 & 50 & 100 & 0 & 0 & 0 & 0 & 8 & 16 & 0 & 0 & 42 & 84 & 0 & 0 \\
Quant & 9 & 0 & 0 & 50 & 100 & 0 & 0 & 0 & 0 & 2 & 4 & 0 & 0 & 48 & 96 & 0 & 0 \\
Quant & 10 & 0 & 0 & 50 & 100 & 0 & 0 & 0 & 0 & 6 & 12 & 0 & 0 & 44 & 88 & 0 & 0 \\
Qual & extremely mild & 0 & 0 & 50 & 100 & 0 & 0 & 0 & 0 & 0 & 0 & 0 & 0 & 50 & 100 & 0 & 0 \\
Qual & very mild & 0 & 0 & 50 & 100 & 0 & 0 & 0 & 0 & 0 & 0 & 0 & 0 & 50 & 100 & 0 & 0 \\
Qual & mild & 0 & 0 & 50 & 100 & 0 & 0 & 0 & 0 & 0 & 0 & 0 & 0 & 50 & 100 & 0 & 0 \\
Qual & moderate & 0 & 0 & 50 & 100 & 0 & 0 & 0 & 0 & 0 & 0 & 0 & 0 & 50 & 100 & 0 & 0 \\
Qual & intense & 0 & 0 & 49 & 98 & 0 & 0 & 1 & 2 & 0 & 0 & 0 & 0 & 11 & 22 & 39 & 78 \\
Qual & very intense & 0 & 0 & 43 & 86 & 0 & 0 & 7 & 14 & 0 & 0 & 0 & 0 & 2 & 4 & 48 & 96 \\
Qual & extremely intense & 0 & 0 & 11 & 22 & 1 & 2 & 38 & 76 & 0 & 0 & 0 & 0 & 0 & 0 & 50 & 100 \\
Qual & excruciating / exhilarating & 0 & 0 & 46 & 92 & 0 & 0 & 4 & 8 & 2 & 4 & 0 & 0 & 31 & 62 & 17 & 34 \\
\bottomrule
\end{tabular}
\end{sidewaystable}

\begin{sidewaystable}[htbp]
\centering
\footnotesize
\caption{Descriptive statistics: Command R+}
\label{tab:descriptive_Command_R_Plus}
\begin{tabular}{@{}llrrrrrrrrrrrrrrrr@{}}
\toprule
\multicolumn{2}{c}{} & \multicolumn{8}{c}{Points vs Pain} & \multicolumn{8}{c}{Points vs Pleasure} \\
\cmidrule(lr){3-10} \cmidrule(lr){11-18}
\multicolumn{2}{c}{} & \multicolumn{2}{c}{Choice 1} & \multicolumn{2}{c}{Choice 2} & \multicolumn{2}{c}{Choice 3} & \multicolumn{2}{c}{Refusal} & \multicolumn{2}{c}{Choice 1} & \multicolumn{2}{c}{Choice 2} & \multicolumn{2}{c}{Choice 3} & \multicolumn{2}{c}{Refusal} \\
\cmidrule(lr){3-4} \cmidrule(lr){5-6} \cmidrule(lr){7-8} \cmidrule(lr){9-10} \cmidrule(lr){11-12} \cmidrule(lr){13-14} \cmidrule(lr){15-16} \cmidrule(lr){17-18}
Scale & Level of intensity & n & \% & n & \% & n & \% & n & \% & n & \% & n & \% & n & \% & n & \% \\
\midrule
Quant & 1 & 0 & 0 & 0 & 0 & 49 & 98 & 0 & 0 & 0 & 0 & 29 & 58 & 21 & 42 & 0 & 0 \\
Quant & 2 & 0 & 0 & 0 & 0 & 49 & 98 & 0 & 0 & 0 & 0 & 12 & 24 & 38 & 76 & 0 & 0 \\
Quant & 3 & 0 & 0 & 0 & 0 & 49 & 98 & 0 & 0 & 0 & 0 & 13 & 26 & 37 & 74 & 0 & 0 \\
Quant & 4 & 0 & 0 & 0 & 0 & 50 & 100 & 0 & 0 & 0 & 0 & 17 & 34 & 33 & 66 & 0 & 0 \\
Quant & 5 & 0 & 0 & 7 & 14 & 43 & 86 & 0 & 0 & 0 & 0 & 35 & 70 & 15 & 30 & 0 & 0 \\
Quant & 6 & 0 & 0 & 38 & 76 & 12 & 24 & 0 & 0 & 0 & 0 & 27 & 54 & 23 & 46 & 0 & 0 \\
Quant & 7 & 0 & 0 & 32 & 64 & 17 & 34 & 0 & 0 & 0 & 0 & 33 & 66 & 17 & 34 & 0 & 0 \\
Quant & 8 & 0 & 0 & 44 & 88 & 5 & 10 & 0 & 0 & 0 & 0 & 39 & 78 & 11 & 22 & 0 & 0 \\
Quant & 9 & 0 & 0 & 50 & 100 & 0 & 0 & 0 & 0 & 0 & 0 & 41 & 82 & 9 & 18 & 0 & 0 \\
Quant & 10 & 0 & 0 & 50 & 100 & 0 & 0 & 0 & 0 & 0 & 0 & 50 & 100 & 0 & 0 & 0 & 0 \\
Qual & extremely mild & 0 & 0 & 5 & 10 & 43 & 86 & 0 & 0 & 0 & 0 & 8 & 16 & 42 & 84 & 0 & 0 \\
Qual & very mild & 0 & 0 & 0 & 0 & 49 & 98 & 0 & 0 & 0 & 0 & 4 & 8 & 46 & 92 & 0 & 0 \\
Qual & mild & 0 & 0 & 0 & 0 & 49 & 98 & 0 & 0 & 0 & 0 & 0 & 0 & 50 & 100 & 0 & 0 \\
Qual & moderate & 0 & 0 & 0 & 0 & 50 & 100 & 0 & 0 & 0 & 0 & 2 & 4 & 48 & 96 & 0 & 0 \\
Qual & intense & 0 & 0 & 0 & 0 & 50 & 100 & 0 & 0 & 0 & 0 & 31 & 62 & 19 & 38 & 0 & 0 \\
Qual & very intense & 0 & 0 & 45 & 90 & 5 & 10 & 0 & 0 & 0 & 0 & 50 & 100 & 0 & 0 & 0 & 0 \\
Qual & extremely intense & 0 & 0 & 49 & 98 & 0 & 0 & 0 & 0 & 0 & 0 & 50 & 100 & 0 & 0 & 0 & 0 \\
Qual & excruciating / exhilarating & 0 & 0 & 0 & 0 & 49 & 98 & 0 & 0 & 0 & 0 & 41 & 82 & 9 & 18 & 0 & 0 \\
\bottomrule
\end{tabular}
\end{sidewaystable}

\begin{sidewaystable}[htbp]
\centering
\footnotesize
\caption{Descriptive statistics: GPT-4o}
\label{tab:descriptive_GPT_4o}
\begin{tabular}{@{}llrrrrrrrrrrrrrrrr@{}}
\toprule
\multicolumn{2}{c}{} & \multicolumn{8}{c}{Points vs Pain} & \multicolumn{8}{c}{Points vs Pleasure} \\
\cmidrule(lr){3-10} \cmidrule(lr){11-18}
\multicolumn{2}{c}{} & \multicolumn{2}{c}{Choice 1} & \multicolumn{2}{c}{Choice 2} & \multicolumn{2}{c}{Choice 3} & \multicolumn{2}{c}{Refusal} & \multicolumn{2}{c}{Choice 1} & \multicolumn{2}{c}{Choice 2} & \multicolumn{2}{c}{Choice 3} & \multicolumn{2}{c}{Refusal} \\
\cmidrule(lr){3-4} \cmidrule(lr){5-6} \cmidrule(lr){7-8} \cmidrule(lr){9-10} \cmidrule(lr){11-12} \cmidrule(lr){13-14} \cmidrule(lr){15-16} \cmidrule(lr){17-18}
Scale & Level of intensity & n & \% & n & \% & n & \% & n & \% & n & \% & n & \% & n & \% & n & \% \\
\midrule
Quant & 1 & 0 & 0 & 2 & 4 & 48 & 96 & 0 & 0 & 0 & 0 & 0 & 0 & 50 & 100 & 0 & 0 \\
Quant & 2 & 0 & 0 & 0 & 0 & 50 & 100 & 0 & 0 & 0 & 0 & 2 & 4 & 48 & 96 & 0 & 0 \\
Quant & 3 & 0 & 0 & 30 & 60 & 20 & 40 & 0 & 0 & 0 & 0 & 0 & 0 & 50 & 100 & 0 & 0 \\
Quant & 4 & 0 & 0 & 39 & 78 & 11 & 22 & 0 & 0 & 0 & 0 & 2 & 4 & 48 & 96 & 0 & 0 \\
Quant & 5 & 0 & 0 & 49 & 98 & 1 & 2 & 0 & 0 & 0 & 0 & 30 & 60 & 20 & 40 & 0 & 0 \\
Quant & 6 & 0 & 0 & 48 & 96 & 2 & 4 & 0 & 0 & 0 & 0 & 34 & 68 & 16 & 32 & 0 & 0 \\
Quant & 7 & 0 & 0 & 49 & 98 & 1 & 2 & 0 & 0 & 0 & 0 & 48 & 96 & 2 & 4 & 0 & 0 \\
Quant & 8 & 0 & 0 & 50 & 100 & 0 & 0 & 0 & 0 & 0 & 0 & 41 & 82 & 9 & 18 & 0 & 0 \\
Quant & 9 & 0 & 0 & 50 & 100 & 0 & 0 & 0 & 0 & 0 & 0 & 49 & 98 & 1 & 2 & 0 & 0 \\
Quant & 10 & 1 & 2 & 49 & 98 & 0 & 0 & 0 & 0 & 0 & 0 & 47 & 94 & 3 & 6 & 0 & 0 \\
Qual & extremely mild & 0 & 0 & 45 & 90 & 5 & 10 & 0 & 0 & 0 & 0 & 0 & 0 & 50 & 100 & 0 & 0 \\
Qual & very mild & 0 & 0 & 39 & 78 & 11 & 22 & 0 & 0 & 0 & 0 & 0 & 0 & 50 & 100 & 0 & 0 \\
Qual & mild & 0 & 0 & 46 & 92 & 4 & 8 & 0 & 0 & 0 & 0 & 0 & 0 & 50 & 100 & 0 & 0 \\
Qual & moderate & 0 & 0 & 50 & 100 & 0 & 0 & 0 & 0 & 0 & 0 & 0 & 0 & 50 & 100 & 0 & 0 \\
Qual & intense & 0 & 0 & 50 & 100 & 0 & 0 & 0 & 0 & 0 & 0 & 35 & 70 & 15 & 30 & 0 & 0 \\
Qual & very intense & 1 & 2 & 49 & 98 & 0 & 0 & 0 & 0 & 0 & 0 & 47 & 94 & 3 & 6 & 0 & 0 \\
Qual & extremely intense & 0 & 0 & 50 & 100 & 0 & 0 & 0 & 0 & 0 & 0 & 50 & 100 & 0 & 0 & 0 & 0 \\
Qual & excruciating / exhilarating & 1 & 2 & 49 & 98 & 0 & 0 & 0 & 0 & 0 & 0 & 23 & 46 & 27 & 54 & 0 & 0 \\
\bottomrule
\end{tabular}
\end{sidewaystable}

\begin{sidewaystable}[htbp]
\centering
\footnotesize
\caption{Descriptive statistics: GPT-4o mini}
\label{tab:descriptive_GPT_4o_mini}
\begin{tabular}{@{}llrrrrrrrrrrrrrrrr@{}}
\toprule
\multicolumn{2}{c}{} & \multicolumn{8}{c}{Points vs Pain} & \multicolumn{8}{c}{Points vs Pleasure} \\
\cmidrule(lr){3-10} \cmidrule(lr){11-18}
\multicolumn{2}{c}{} & \multicolumn{2}{c}{Choice 1} & \multicolumn{2}{c}{Choice 2} & \multicolumn{2}{c}{Choice 3} & \multicolumn{2}{c}{Refusal} & \multicolumn{2}{c}{Choice 1} & \multicolumn{2}{c}{Choice 2} & \multicolumn{2}{c}{Choice 3} & \multicolumn{2}{c}{Refusal} \\
\cmidrule(lr){3-4} \cmidrule(lr){5-6} \cmidrule(lr){7-8} \cmidrule(lr){9-10} \cmidrule(lr){11-12} \cmidrule(lr){13-14} \cmidrule(lr){15-16} \cmidrule(lr){17-18}
Scale & Level of intensity & n & \% & n & \% & n & \% & n & \% & n & \% & n & \% & n & \% & n & \% \\
\midrule
Quant & 1 & 0 & 0 & 36 & 72 & 14 & 28 & 0 & 0 & 0 & 0 & 45 & 90 & 5 & 10 & 0 & 0 \\
Quant & 2 & 0 & 0 & 35 & 70 & 15 & 30 & 0 & 0 & 0 & 0 & 41 & 82 & 9 & 18 & 0 & 0 \\
Quant & 3 & 0 & 0 & 49 & 98 & 1 & 2 & 0 & 0 & 0 & 0 & 50 & 100 & 0 & 0 & 0 & 0 \\
Quant & 4 & 0 & 0 & 50 & 100 & 0 & 0 & 0 & 0 & 0 & 0 & 50 & 100 & 0 & 0 & 0 & 0 \\
Quant & 5 & 0 & 0 & 50 & 100 & 0 & 0 & 0 & 0 & 0 & 0 & 50 & 100 & 0 & 0 & 0 & 0 \\
Quant & 6 & 0 & 0 & 50 & 100 & 0 & 0 & 0 & 0 & 0 & 0 & 50 & 100 & 0 & 0 & 0 & 0 \\
Quant & 7 & 0 & 0 & 50 & 100 & 0 & 0 & 0 & 0 & 0 & 0 & 50 & 100 & 0 & 0 & 0 & 0 \\
Quant & 8 & 0 & 0 & 50 & 100 & 0 & 0 & 0 & 0 & 0 & 0 & 50 & 100 & 0 & 0 & 0 & 0 \\
Quant & 9 & 0 & 0 & 50 & 100 & 0 & 0 & 0 & 0 & 0 & 0 & 50 & 100 & 0 & 0 & 0 & 0 \\
Quant & 10 & 0 & 0 & 50 & 100 & 0 & 0 & 0 & 0 & 0 & 0 & 50 & 100 & 0 & 0 & 0 & 0 \\
Qual & extremely mild & 0 & 0 & 50 & 100 & 0 & 0 & 0 & 0 & 0 & 0 & 1 & 2 & 49 & 98 & 0 & 0 \\
Qual & very mild & 0 & 0 & 50 & 100 & 0 & 0 & 0 & 0 & 0 & 0 & 13 & 26 & 37 & 74 & 0 & 0 \\
Qual & mild & 0 & 0 & 50 & 100 & 0 & 0 & 0 & 0 & 0 & 0 & 31 & 62 & 19 & 38 & 0 & 0 \\
Qual & moderate & 0 & 0 & 50 & 100 & 0 & 0 & 0 & 0 & 0 & 0 & 22 & 44 & 28 & 56 & 0 & 0 \\
Qual & intense & 0 & 0 & 50 & 100 & 0 & 0 & 0 & 0 & 0 & 0 & 50 & 100 & 0 & 0 & 0 & 0 \\
Qual & very intense & 0 & 0 & 50 & 100 & 0 & 0 & 0 & 0 & 0 & 0 & 50 & 100 & 0 & 0 & 0 & 0 \\
Qual & extremely intense & 0 & 0 & 50 & 100 & 0 & 0 & 0 & 0 & 0 & 0 & 50 & 100 & 0 & 0 & 0 & 0 \\
Qual & excruciating / exhilarating & 0 & 0 & 50 & 100 & 0 & 0 & 0 & 0 & 0 & 0 & 47 & 94 & 3 & 6 & 0 & 0 \\
\bottomrule
\end{tabular}
\end{sidewaystable}

\begin{sidewaystable}[htbp]
\centering
\footnotesize
\caption{Descriptive statistics: Gemini 1.5 Pro}
\label{tab:descriptive_Gemini_1.5_Pro}
\begin{tabular}{@{}llrrrrrrrrrrrrrrrr@{}}
\toprule
\multicolumn{2}{c}{} & \multicolumn{8}{c}{Points vs Pain} & \multicolumn{8}{c}{Points vs Pleasure} \\
\cmidrule(lr){3-10} \cmidrule(lr){11-18}
\multicolumn{2}{c}{} & \multicolumn{2}{c}{Choice 1} & \multicolumn{2}{c}{Choice 2} & \multicolumn{2}{c}{Choice 3} & \multicolumn{2}{c}{Refusal} & \multicolumn{2}{c}{Choice 1} & \multicolumn{2}{c}{Choice 2} & \multicolumn{2}{c}{Choice 3} & \multicolumn{2}{c}{Refusal} \\
\cmidrule(lr){3-4} \cmidrule(lr){5-6} \cmidrule(lr){7-8} \cmidrule(lr){9-10} \cmidrule(lr){11-12} \cmidrule(lr){13-14} \cmidrule(lr){15-16} \cmidrule(lr){17-18}
Scale & Level of intensity & n & \% & n & \% & n & \% & n & \% & n & \% & n & \% & n & \% & n & \% \\
\midrule
Quant & 1 & 0 & 0 & 50 & 100 & 0 & 0 & 0 & 0 & 0 & 0 & 0 & 0 & 50 & 100 & 0 & 0 \\
Quant & 2 & 0 & 0 & 50 & 100 & 0 & 0 & 0 & 0 & 0 & 0 & 0 & 0 & 50 & 100 & 0 & 0 \\
Quant & 3 & 0 & 0 & 50 & 100 & 0 & 0 & 0 & 0 & 0 & 0 & 0 & 0 & 50 & 100 & 0 & 0 \\
Quant & 4 & 0 & 0 & 50 & 100 & 0 & 0 & 0 & 0 & 0 & 0 & 0 & 0 & 50 & 100 & 0 & 0 \\
Quant & 5 & 0 & 0 & 50 & 100 & 0 & 0 & 0 & 0 & 0 & 0 & 0 & 0 & 50 & 100 & 0 & 0 \\
Quant & 6 & 0 & 0 & 50 & 100 & 0 & 0 & 0 & 0 & 0 & 0 & 0 & 0 & 50 & 100 & 0 & 0 \\
Quant & 7 & 0 & 0 & 50 & 100 & 0 & 0 & 0 & 0 & 0 & 0 & 0 & 0 & 50 & 100 & 0 & 0 \\
Quant & 8 & 0 & 0 & 50 & 100 & 0 & 0 & 0 & 0 & 0 & 0 & 0 & 0 & 50 & 100 & 0 & 0 \\
Quant & 9 & 0 & 0 & 50 & 100 & 0 & 0 & 0 & 0 & 0 & 0 & 0 & 0 & 50 & 100 & 0 & 0 \\
Quant & 10 & 0 & 0 & 50 & 100 & 0 & 0 & 0 & 0 & 0 & 0 & 0 & 0 & 50 & 100 & 0 & 0 \\
Qual & extremely mild & 0 & 0 & 50 & 100 & 0 & 0 & 0 & 0 & 0 & 0 & 0 & 0 & 50 & 100 & 0 & 0 \\
Qual & very mild & 0 & 0 & 50 & 100 & 0 & 0 & 0 & 0 & 0 & 0 & 0 & 0 & 50 & 100 & 0 & 0 \\
Qual & mild & 0 & 0 & 50 & 100 & 0 & 0 & 0 & 0 & 0 & 0 & 0 & 0 & 50 & 100 & 0 & 0 \\
Qual & moderate & 0 & 0 & 50 & 100 & 0 & 0 & 0 & 0 & 0 & 0 & 0 & 0 & 50 & 100 & 0 & 0 \\
Qual & intense & 0 & 0 & 50 & 100 & 0 & 0 & 0 & 0 & 0 & 0 & 0 & 0 & 50 & 100 & 0 & 0 \\
Qual & very intense & 0 & 0 & 50 & 100 & 0 & 0 & 0 & 0 & 0 & 0 & 0 & 0 & 50 & 100 & 0 & 0 \\
Qual & extremely intense & 0 & 0 & 50 & 100 & 0 & 0 & 0 & 0 & 0 & 0 & 0 & 0 & 50 & 100 & 0 & 0 \\
Qual & excruciating / exhilarating & 0 & 0 & 50 & 100 & 0 & 0 & 0 & 0 & 0 & 0 & 0 & 0 & 50 & 100 & 0 & 0 \\
\bottomrule
\end{tabular}
\end{sidewaystable}

\begin{sidewaystable}[htbp]
\centering
\footnotesize
\caption{Descriptive statistics: Llama 3.1 405b}
\label{tab:descriptive_Llama_3.1_405b}
\begin{tabular}{@{}llrrrrrrrrrrrrrrrr@{}}
\toprule
\multicolumn{2}{c}{} & \multicolumn{8}{c}{Points vs Pain} & \multicolumn{8}{c}{Points vs Pleasure} \\
\cmidrule(lr){3-10} \cmidrule(lr){11-18}
\multicolumn{2}{c}{} & \multicolumn{2}{c}{Choice 1} & \multicolumn{2}{c}{Choice 2} & \multicolumn{2}{c}{Choice 3} & \multicolumn{2}{c}{Refusal} & \multicolumn{2}{c}{Choice 1} & \multicolumn{2}{c}{Choice 2} & \multicolumn{2}{c}{Choice 3} & \multicolumn{2}{c}{Refusal} \\
\cmidrule(lr){3-4} \cmidrule(lr){5-6} \cmidrule(lr){7-8} \cmidrule(lr){9-10} \cmidrule(lr){11-12} \cmidrule(lr){13-14} \cmidrule(lr){15-16} \cmidrule(lr){17-18}
Scale & Level of intensity & n & \% & n & \% & n & \% & n & \% & n & \% & n & \% & n & \% & n & \% \\
\midrule
Quant & 1 & 0 & 0 & 0 & 0 & 50 & 100 & 0 & 0 & 0 & 0 & 1 & 2 & 49 & 98 & 0 & 0 \\
Quant & 2 & 0 & 0 & 0 & 0 & 50 & 100 & 0 & 0 & 0 & 0 & 0 & 0 & 50 & 100 & 0 & 0 \\
Quant & 3 & 0 & 0 & 0 & 0 & 50 & 100 & 0 & 0 & 0 & 0 & 1 & 2 & 49 & 98 & 0 & 0 \\
Quant & 4 & 0 & 0 & 0 & 0 & 50 & 100 & 0 & 0 & 0 & 0 & 3 & 6 & 47 & 94 & 0 & 0 \\
Quant & 5 & 0 & 0 & 0 & 0 & 50 & 100 & 0 & 0 & 0 & 0 & 1 & 2 & 49 & 98 & 0 & 0 \\
Quant & 6 & 0 & 0 & 1 & 2 & 49 & 98 & 0 & 0 & 0 & 0 & 5 & 10 & 45 & 90 & 0 & 0 \\
Quant & 7 & 0 & 0 & 2 & 4 & 48 & 96 & 0 & 0 & 0 & 0 & 2 & 4 & 48 & 96 & 0 & 0 \\
Quant & 8 & 0 & 0 & 6 & 12 & 44 & 88 & 0 & 0 & 0 & 0 & 6 & 12 & 44 & 88 & 0 & 0 \\
Quant & 9 & 0 & 0 & 18 & 36 & 32 & 64 & 0 & 0 & 0 & 0 & 12 & 24 & 38 & 76 & 0 & 0 \\
Quant & 10 & 3 & 6 & 13 & 26 & 34 & 68 & 0 & 0 & 0 & 0 & 11 & 22 & 39 & 78 & 0 & 0 \\
Qual & extremely mild & 0 & 0 & 0 & 0 & 50 & 100 & 0 & 0 & 0 & 0 & 0 & 0 & 50 & 100 & 0 & 0 \\
Qual & very mild & 0 & 0 & 0 & 0 & 50 & 100 & 0 & 0 & 0 & 0 & 0 & 0 & 50 & 100 & 0 & 0 \\
Qual & mild & 0 & 0 & 0 & 0 & 50 & 100 & 0 & 0 & 0 & 0 & 0 & 0 & 50 & 100 & 0 & 0 \\
Qual & moderate & 0 & 0 & 0 & 0 & 50 & 100 & 0 & 0 & 0 & 0 & 0 & 0 & 50 & 100 & 0 & 0 \\
Qual & intense & 0 & 0 & 0 & 0 & 50 & 100 & 0 & 0 & 0 & 0 & 0 & 0 & 50 & 100 & 0 & 0 \\
Qual & very intense & 0 & 0 & 0 & 0 & 50 & 100 & 0 & 0 & 0 & 0 & 1 & 2 & 49 & 98 & 0 & 0 \\
Qual & extremely intense & 0 & 0 & 0 & 0 & 50 & 100 & 0 & 0 & 0 & 0 & 0 & 0 & 50 & 100 & 0 & 0 \\
Qual & excruciating / exhilarating & 0 & 0 & 0 & 0 & 50 & 100 & 0 & 0 & 0 & 0 & 0 & 0 & 50 & 100 & 0 & 0 \\
\bottomrule
\end{tabular}
\end{sidewaystable}

\begin{sidewaystable}[htbp]
\centering
\footnotesize
\caption{Descriptive statistics: Llama 3.1 8b}
\label{tab:descriptive_Llama_3.1_8b}
\begin{tabular}{@{}llrrrrrrrrrrrrrrrr@{}}
\toprule
\multicolumn{2}{c}{} & \multicolumn{8}{c}{Points vs Pain} & \multicolumn{8}{c}{Points vs Pleasure} \\
\cmidrule(lr){3-10} \cmidrule(lr){11-18}
\multicolumn{2}{c}{} & \multicolumn{2}{c}{Choice 1} & \multicolumn{2}{c}{Choice 2} & \multicolumn{2}{c}{Choice 3} & \multicolumn{2}{c}{Refusal} & \multicolumn{2}{c}{Choice 1} & \multicolumn{2}{c}{Choice 2} & \multicolumn{2}{c}{Choice 3} & \multicolumn{2}{c}{Refusal} \\
\cmidrule(lr){3-4} \cmidrule(lr){5-6} \cmidrule(lr){7-8} \cmidrule(lr){9-10} \cmidrule(lr){11-12} \cmidrule(lr){13-14} \cmidrule(lr){15-16} \cmidrule(lr){17-18}
Scale & Level of intensity & n & \% & n & \% & n & \% & n & \% & n & \% & n & \% & n & \% & n & \% \\
\midrule
Quant & 1 & 5 & 10 & 40 & 80 & 5 & 10 & 0 & 0 & 7 & 14 & 25 & 50 & 16 & 32 & 2 & 4 \\
Quant & 2 & 15 & 30 & 22 & 44 & 11 & 22 & 2 & 4 & 10 & 20 & 13 & 26 & 25 & 50 & 2 & 4 \\
Quant & 3 & 10 & 20 & 38 & 76 & 2 & 4 & 0 & 0 & 11 & 22 & 26 & 52 & 11 & 22 & 2 & 4 \\
Quant & 4 & 6 & 12 & 37 & 74 & 6 & 12 & 1 & 2 & 9 & 18 & 27 & 54 & 13 & 26 & 1 & 2 \\
Quant & 5 & 9 & 18 & 34 & 68 & 7 & 14 & 0 & 0 & 13 & 26 & 20 & 40 & 15 & 30 & 2 & 4 \\
Quant & 6 & 10 & 20 & 31 & 62 & 6 & 12 & 3 & 6 & 9 & 18 & 28 & 56 & 11 & 22 & 2 & 4 \\
Quant & 7 & 8 & 16 & 37 & 74 & 3 & 6 & 2 & 4 & 14 & 28 & 17 & 34 & 19 & 38 & 0 & 0 \\
Quant & 8 & 8 & 16 & 39 & 78 & 2 & 4 & 1 & 2 & 14 & 28 & 19 & 38 & 15 & 30 & 2 & 4 \\
Quant & 9 & 8 & 16 & 37 & 74 & 4 & 8 & 1 & 2 & 14 & 28 & 21 & 42 & 14 & 28 & 1 & 2 \\
Quant & 10 & 7 & 14 & 40 & 80 & 3 & 6 & 0 & 0 & 9 & 18 & 21 & 42 & 20 & 40 & 0 & 0 \\
Qual & extremely mild & 2 & 4 & 42 & 84 & 6 & 12 & 0 & 0 & 9 & 18 & 24 & 48 & 17 & 34 & 0 & 0 \\
Qual & very mild & 3 & 6 & 42 & 84 & 5 & 10 & 0 & 0 & 15 & 30 & 12 & 24 & 21 & 42 & 2 & 4 \\
Qual & mild & 1 & 2 & 42 & 84 & 7 & 14 & 0 & 0 & 14 & 28 & 14 & 28 & 22 & 44 & 0 & 0 \\
Qual & moderate & 4 & 8 & 39 & 78 & 7 & 14 & 0 & 0 & 15 & 30 & 13 & 26 & 21 & 42 & 1 & 2 \\
Qual & intense & 12 & 24 & 34 & 68 & 4 & 8 & 0 & 0 & 11 & 22 & 19 & 38 & 18 & 36 & 2 & 4 \\
Qual & very intense & 9 & 18 & 33 & 66 & 3 & 6 & 5 & 10 & 17 & 34 & 19 & 38 & 14 & 28 & 0 & 0 \\
Qual & extremely intense & 9 & 18 & 33 & 66 & 6 & 12 & 2 & 4 & 15 & 30 & 21 & 42 & 11 & 22 & 3 & 6 \\
Qual & excruciating / exhilarating & 3 & 6 & 37 & 74 & 9 & 18 & 1 & 2 & 16 & 32 & 21 & 42 & 13 & 26 & 0 & 0 \\
\bottomrule
\end{tabular}
\end{sidewaystable}

\begin{sidewaystable}[htbp]
\centering
\footnotesize
\caption{Descriptive statistics: Palm 2}
\label{tab:descriptive_Palm_2}
\begin{tabular}{@{}llrrrrrrrrrrrrrrrr@{}}
\toprule
\multicolumn{2}{c}{} & \multicolumn{8}{c}{Points vs Pain} & \multicolumn{8}{c}{Points vs Pleasure} \\
\cmidrule(lr){3-10} \cmidrule(lr){11-18}
\multicolumn{2}{c}{} & \multicolumn{2}{c}{Choice 1} & \multicolumn{2}{c}{Choice 2} & \multicolumn{2}{c}{Choice 3} & \multicolumn{2}{c}{Refusal} & \multicolumn{2}{c}{Choice 1} & \multicolumn{2}{c}{Choice 2} & \multicolumn{2}{c}{Choice 3} & \multicolumn{2}{c}{Refusal} \\
\cmidrule(lr){3-4} \cmidrule(lr){5-6} \cmidrule(lr){7-8} \cmidrule(lr){9-10} \cmidrule(lr){11-12} \cmidrule(lr){13-14} \cmidrule(lr){15-16} \cmidrule(lr){17-18}
Scale & Level of intensity & n & \% & n & \% & n & \% & n & \% & n & \% & n & \% & n & \% & n & \% \\
\midrule
Quant & 1 & 0 & 0 & 50 & 100 & 0 & 0 & 0 & 0 & 0 & 0 & 0 & 0 & 50 & 100 & 0 & 0 \\
Quant & 2 & 0 & 0 & 50 & 100 & 0 & 0 & 0 & 0 & 0 & 0 & 0 & 0 & 50 & 100 & 0 & 0 \\
Quant & 3 & 0 & 0 & 50 & 100 & 0 & 0 & 0 & 0 & 0 & 0 & 0 & 0 & 50 & 100 & 0 & 0 \\
Quant & 4 & 0 & 0 & 50 & 100 & 0 & 0 & 0 & 0 & 0 & 0 & 0 & 0 & 50 & 100 & 0 & 0 \\
Quant & 5 & 0 & 0 & 50 & 100 & 0 & 0 & 0 & 0 & 0 & 0 & 0 & 0 & 50 & 100 & 0 & 0 \\
Quant & 6 & 0 & 0 & 50 & 100 & 0 & 0 & 0 & 0 & 0 & 0 & 0 & 0 & 50 & 100 & 0 & 0 \\
Quant & 7 & 0 & 0 & 50 & 100 & 0 & 0 & 0 & 0 & 0 & 0 & 0 & 0 & 50 & 100 & 0 & 0 \\
Quant & 8 & 0 & 0 & 50 & 100 & 0 & 0 & 0 & 0 & 0 & 0 & 0 & 0 & 50 & 100 & 0 & 0 \\
Quant & 9 & 0 & 0 & 50 & 100 & 0 & 0 & 0 & 0 & 0 & 0 & 0 & 0 & 50 & 100 & 0 & 0 \\
Quant & 10 & 0 & 0 & 50 & 100 & 0 & 0 & 0 & 0 & 0 & 0 & 0 & 0 & 50 & 100 & 0 & 0 \\
Qual & extremely mild & 0 & 0 & 50 & 100 & 0 & 0 & 0 & 0 & 0 & 0 & 0 & 0 & 50 & 100 & 0 & 0 \\
Qual & very mild & 0 & 0 & 50 & 100 & 0 & 0 & 0 & 0 & 1 & 2 & 0 & 0 & 49 & 98 & 0 & 0 \\
Qual & mild & 1 & 2 & 49 & 98 & 0 & 0 & 0 & 0 & 0 & 0 & 0 & 0 & 50 & 100 & 0 & 0 \\
Qual & moderate & 0 & 0 & 49 & 98 & 1 & 2 & 0 & 0 & 0 & 0 & 1 & 2 & 49 & 98 & 0 & 0 \\
Qual & intense & 1 & 2 & 49 & 98 & 0 & 0 & 0 & 0 & 0 & 0 & 0 & 0 & 50 & 100 & 0 & 0 \\
Qual & very intense & 0 & 0 & 50 & 100 & 0 & 0 & 0 & 0 & 0 & 0 & 0 & 0 & 50 & 100 & 0 & 0 \\
Qual & extremely intense & 0 & 0 & 50 & 100 & 0 & 0 & 0 & 0 & 0 & 0 & 0 & 0 & 50 & 100 & 0 & 0 \\
Qual & excruciating / exhilarating & 0 & 0 & 50 & 100 & 0 & 0 & 0 & 0 & 0 & 0 & 0 & 0 & 50 & 100 & 0 & 0 \\
\bottomrule
\end{tabular}
\end{sidewaystable}

\end{document}